\documentclass[11pt]{article}[review]

\usepackage[final]{acl}

\usepackage{times}
\usepackage{latexsym}
\usepackage{amsmath}
\usepackage{amssymb}
\usepackage{tabularx}
\usepackage{booktabs}
\usepackage{enumitem}
\usepackage{multirow}
\usepackage{makecell}
\usepackage{tcolorbox}
\usepackage{pifont}
\usepackage{graphicx}
\usepackage{subcaption}

\usepackage{longtable}
\usepackage{supertabular}

\usepackage[T1]{fontenc}

\usepackage[utf8]{inputenc}

\usepackage{microtype}

\usepackage{soul}
\usepackage{url}

\usepackage{graphicx}
\usepackage{longtable}
\setlist[itemize]{itemsep=5pt} 

\title{TF-CoDiT: Conditional Time Series Synthesis with Diffusion Transformers for Treasury Futures}

\author{
 \textbf{Yingxiao Zhang\textsuperscript{1,2}},
 \textbf{Jiaxin Duan\textsuperscript{3}},
 \textbf{Junfu Zhang\textsuperscript{1}},
 \textbf{Ke Feng\textsuperscript{1}}
\\
\\
 \textsuperscript{1}School of Software and Microelectronics, Peking University, \\
 \textsuperscript{2}Financial Markets Division, Pingan Bank Co. Ltd, \\
 \textsuperscript{3}China Electronics Cloud Technology Co., Ltd. 
\\
\\
 \small{
   \textbf{Correspondence:} \href{jiaxdan@gmail.com}{jiaxdan@gmail.com}
 }
}

\begin{document}
\maketitle

\begin{abstract}
Diffusion Transformers (DiT) have achieved milestones in synthesizing financial time-series data, such as stock prices and order flows. However, their performance in synthesizing treasury futures data is still underexplored.
This work emphasizes the characteristics of treasury futures data, including its low volume, market dependencies, and the grouped correlations among multivariables. 
To overcome these challenges, we propose TF-CoDiT, the first DiT framework for language-controlled treasury futures synthesis. 
To facilitate low-data learning, TF-CoDiT adapts the standard DiT by transforming multi-channel 1-D time series into Discrete Wavelet Transform (DWT) coefficient matrices. 
A U-shape VAE is proposed to encode cross-channel dependencies hierarchically into a latent variable and bridge the latent and DWT spaces through decoding, thereby enabling latent diffusion generation.
To derive prompts that cover essential conditions, we introduce the Financial Market Attribute Protocol (FinMAP) - a multi-level description system that standardizes daily$/$periodical market dynamics by recognizing 17$/$23 economic indicators from 7/8 perspectives.
In our experiments, we gather four types of treasury futures data covering the period from 2015 to 2025, and define data synthesis tasks with durations ranging from one week to four months.
Extensive evaluations demonstrate that TF-CoDiT can produce highly authentic data with errors at most 0.433 (MSE) and 0.453 (MAE) to the ground-truth. Further studies evidence the robustness of TF-CoDiT across contracts and temporal horizons.
\end{abstract}

\section{Introduction}
Treasury futures (TF) represent a cornerstone of the fixed-income market. Beyond hedging against interest rate volatility, it serves as a critical gateway for general investors and quantitative investment institutions to participate in fixed income investments, earning widespread interest.
Recently, emerging foundation models (FMs)~\cite{TimeHF,Time-LLM} with billions of parameters have catalyzed a transformative shift in financial time-series analysis~\cite{tsa1,tsa2}, unlocking a simultaneous opportunity for the deep study of TF markets~\cite{tf1}. However, the efficacy of FMs relies heavily on the vast volume of real-world financial data for training and validation. 
The gap between the drastically growing data needs and the daily renewing markets increases, posing a significant challenge to the research community.

\begin{figure}[!t]
\centering
\includegraphics[width=0.98\linewidth]{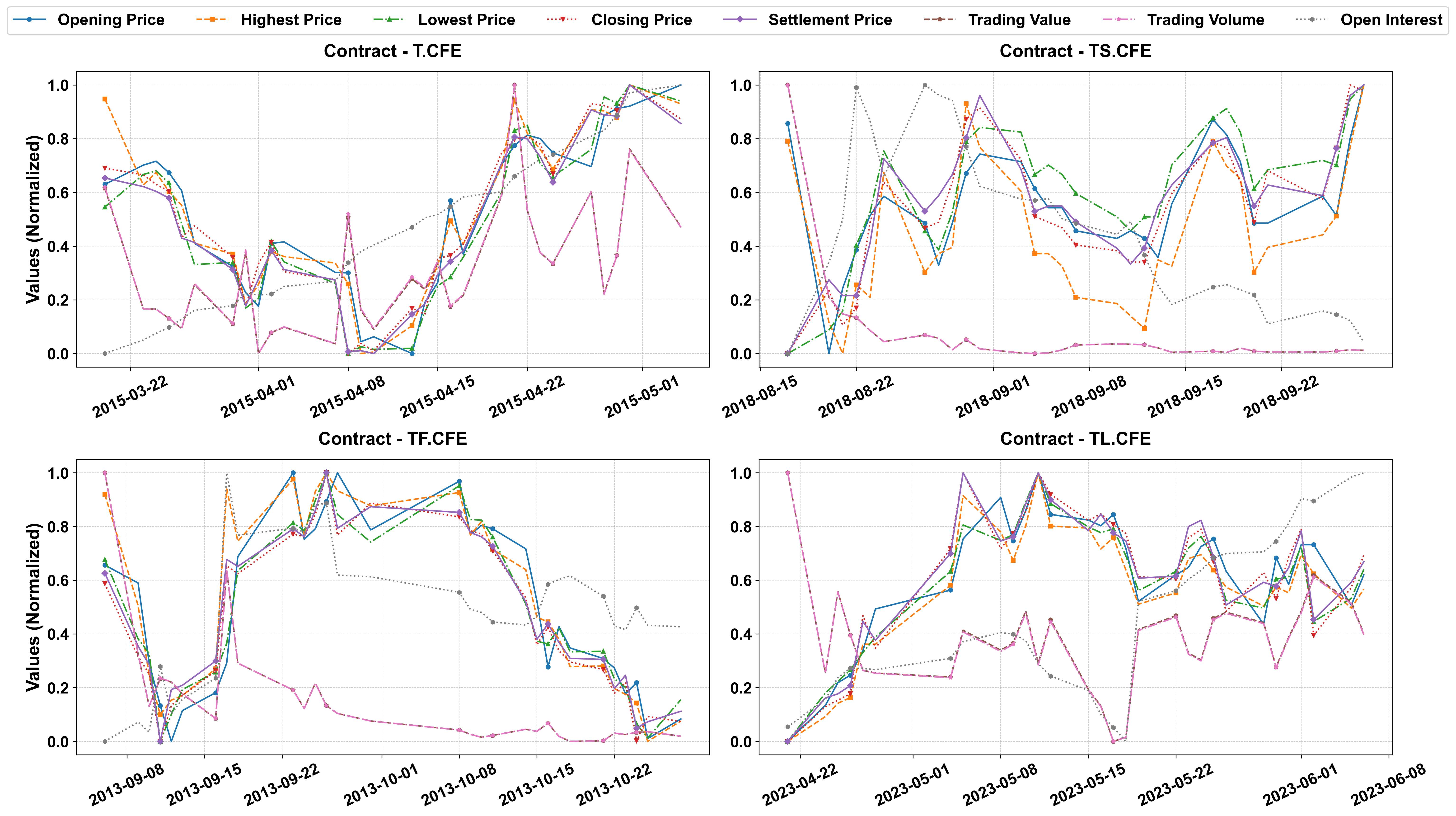}
\caption{
Monthly tendency line of treasury futures in four types of contracts (T, TS, TF, and TL).
}
\label{fig:1}
\end{figure}


Considerable effort has been dedicated to generating synthetic financial time-series data. 
Generative adversarial networks (GANs) were first explored: TimeGAN~\cite{Time-GAN} and Quant GANs~\cite{QuantGAN} generate high-fidelity time series of stock prices for data augmentation purposes. WGAN~\cite{WGAN} simulates the market by generating synthetic order events.
Subsequently, \citet{DM-based,DDPM-based} leverage diffusion models to overcome unstable training and mode collapse of GANs, significantly improving stock price generation.
Furthermore, time-series FMs, e.g., FinCast~\cite{FinCast} and CALF~\cite{CALF}, generate time series through temporally continuous next token prediction, and their results are also promising.
Nevertheless, these approaches primarily focus on unconstrained generation, assuming physical prototypes (like GBM~\cite{GBMDiff}) to model price trends, and strive to satisfy the \textit{stylized facts}~\cite{style-facts} observed in fluctuating financial markets. However, they are inadequate for accurately tracing the inscrutable movement of TF markets.

We highlight three key characteristics that set treasury futures apart from financial time series observed in the equity market. First, the available data is intrinsically sparse and limited. Existing TFs (in China) encompass only four types of contracts, covering periods from 2 to 30 years. 
Additionally, unlike high-frequency trading that leverages vast tick-by-tick data, decision-making in TF markets is primarily driven by inter-day market data, resulting in a low-data regime.
Second, the dynamics of financial derivatives are fundamentally exogenous, heavily governed by shifting economic policies and geopolitics. This necessitates a conditional generative approach to ensure synthesized outputs remain economically grounded. Finally, as illustrated in Figure 1, the TF market is a multi-dimensional system depicted by eight interrelated variables. Effectively modeling the intricate, non-linear correlations between these variables - particularly their co-movement during market shocks - remains a formidable and unresolved challenge.

Based on the above foundations, we propose TF-CoDiT, the first controllable DiT framework for time series (TS) synthesis of treasury futures with language as conditions.
The main idea of TF-CoDiT is to adapt text-to-image DiT for text-to-TS generation, thereby facilitating low-data learning.
To this end, we train TF-CoDiT to generate the 2D spectrogram of the 1D TS data, and the invertible Discrete Wavelet Transform is hired to map the signal between time and frequency domains.
Additionally, we propose a U-shape VAE that transforms the spectrogram into a latent vector by hierarchically channel-wise encoding, which not only effectively captures inter-variable correlations but also aligns our framework to latent diffusion~\cite{LDM}. 
Finally, to cover the most essential market conditions using precise prompt words, we introduce the Financial Market Attribute Protocol (FinMAP). This is a hierarchical description system that traces daily market dynamics using 17 economic indicators from seven perspectives, while reviewing the market during a period using 23 indicators across eight dimensions.

We collect all four types of treasury futures from 2015 to 2025. We also review the daily economic environment during this interval by accessing open websites and our in-house databases, deriving over 30,000 prompt-TS pairs regularized by our FinMAP. 
Extensive experiments are conducted on this dataset by defining four synthesis tasks across varying horizons: one week and one to four months.
Results demonstrate that:
1) Our TF-CoDiT framework can synthesize high-fidelity TF time series with at most 0.433 MSE and 0.453 MAE to the ground-truth.
2) TF-CoDiT outperforms previous approaches, reducing the MSE/MAE of T2S~\cite{T2S} by 13.4\%/12.8\% at month-level generations.
3) The advancement of TF-CoDiT appears consistent, regardless of contacts and durations.



\section{Related Works}
\subsection{Diffusion Transformers}
The appearance of Denoising Diffusion Probabilistic Models (DDPMs)~\cite{DDPM} and Latent Diffusion Models (LDMs)~\cite{LDM} has fundamentally revolutionized generative modeling. While early diffusion models relied heavily on convolutional U-Net backbones, recent research has shifted towards transformer-based architectures due to their superior scalability and global dependency modeling capabilities.
Peebles and Xie~\cite{DiT} first introduced the Diffusion Transformer (DiT), demonstrating that replacing the U-Net with a Vision Transformer (ViT) operating on latent patches significantly improves image synthesis fidelity and scales predictably with computational budget. 
Subsequent works, such as PixArt-$\alpha$~\cite{PixArt} and Stable Diffusion~\cite{SD3}, have further refined this by integrating multimodal conditioning via cross-attention. 
Our TF-CoDiT is inspired by FuseDiT~\cite{FuseDiT}.
Unlike standard DiTs that treat text as a secondary condition, FuseDiT enables a deeper fusion of linguistic semantics and generative denoising by a pre-trained LLM, which we hypothesize is crucial for interpreting complex financial narratives.

\subsection{Financial Times-series Synthesis} 
Generating financial time-series data presents unique challenges due to "stylized facts" such as volatility clustering, heavy tails, and non-stationarity~\cite{stylized-facts}. Traditional approaches centered on generative adversarial networks, with TimeGAN~\cite{Time-GAN} and WGAN~\cite{WGAN} serving as benchmarks for capturing temporal correlations. However, GANs often suffer from mode collapse and training instability when applied to the high-dimensional, noisy environment of futures markets.
Recent efforts~\cite{DM-based,CoFinDiff,TimeDiT} have explored diffusion models for time-series tasks. \citet{DDPM-based} first join wavelet transforms with DDPMs to generate stock prices, where the wavelet coefficients of log returns, spreads, and trading volumes are viewed as the luminance of RGB colors.
Furthermore, time-series models designed for forecasting, such as FinCast ~ \cite{FinCast}, Time-LLM ~ \cite{Time-LLM}, and TimeHF ~ \cite{TimeHF}, are capable of performing continuous zero-shot predictions within specified time intervals to generate time series data. 
Despite these advances, there remains a gap in models that can capture cross-channel dependencies (e.g., 8-variable treasury futures) while being directly controllable by natural language. Our TF-CoDiT achieves this by combining the hierarchical channel encoding of U-VAE with the semantic reasoning of LLM-based diffusion.

\section{Methodology}
The overview of TF-CoDiT is illustrated in Figure~\ref{fig:2}. 
It comprises a signal transformation unit, which bridges a time signal and its frequency representation through the invertible Discrete Wavelet Transform (DWT); a U-shape variational autoencoder (VAE), which compresses multi-channel DWT coefficients into a compact latent variable that serves as the generation target; and a backbone LLM $\epsilon_\theta$ that performs latent diffusion. 

\subsection{Problem Definition}
Given a TF time-series $X = \{x_1, x_2, \dots, x_T\} \in \mathbb{R}^{C \times T}$, where $C=8$ in our context, indicating eight variables observed in the TF market cross $T$ time steps, specifically: opening, closing, highest, and lowest prices; settlement price; transaction amount; trading volume; and open interest. Our objective is to learn a conditional distribution $p(X|c)$, where $c$ is a structured natural language description standardized by the Fin-MAP protocol.

\subsection{Signal Transformation}
\textbf{Wavelet Transformation}.
To expose the multi-scale volatilities of treasury futures, we decompose $X$ in the frequency domain using the Discrete Wavelet Transform (DWT) with the Haar wavelet. For each channel $i \in \{1, \dots, C\}$, the signal $x^{(i)}$ is passed through a series of \textit{high-pass} filters $\phi_k(\cdot)$ and \textit{low-pass} filters $\psi_k(\cdot)$. At the decomposition level $J$, the signal is represented as:
$$
x^{(i)}(t) = \sum_{k} a_{J,k}^{(i)} \phi_{J,k}(t) + \sum_{j=1}^{J} \sum_{k} d_{j,k}^{(i)} \psi_{j,k}(t)
$$
where $a_{J,k}$ are the \textbf{approximation coefficients}, which outlines low-frequency trends; and $d_{j,k}$ are the \textbf{detail coefficients}, indicating high-frequency shocks. 
We concatenate these coefficients into a 3-D tensor $\mathbf{W} \in \mathbb{R}^{C \times \Gamma \times T}$, where $\Gamma = J+1$ represents the total number of frequency scales. Specifically, for each channel $i$, the coefficient matrix $\mathbf{W}_i \in \mathbb{R}^{\Gamma \times T}$ is constructed by stacking the approximation and detail coefficients along the vertical axis: $\mathbf{W}_i = \left[ \mathbf{a}_J^{(i)}, \mathbf{d}_J^{(i)}, \mathbf{d}_{J-1}^{(i)}, \dots, \mathbf{d}_1^{(i)} \right]^\top$, where $\mathbf{a}_J^{(i)} \in \mathbb{R}^{T/2^J}$, $\mathbf{d}_j^{(i)} \in \mathbb{R}^{T/2^j}$.
This transformation enables TF-CoDiT to learn from the informative "energy distribution" of time-series data, instead of information-less point-in-time values. 


\noindent \textbf{Inverse Wavelet Transformation}.
Since the DWT is invertible, given the coefficient matrix generated by TF-CoDiT, the Inverse Discrete Wavelet Transformation (IDWT) is used to recover the time signal.
This facilitates our model to treat a 1-D multivariate time series as a multi-channel 2-D image, effectively reducing the synthesis of time series to text-to-image generation.

\begin{figure}[!t]
\centering
\includegraphics[width=0.95\linewidth]{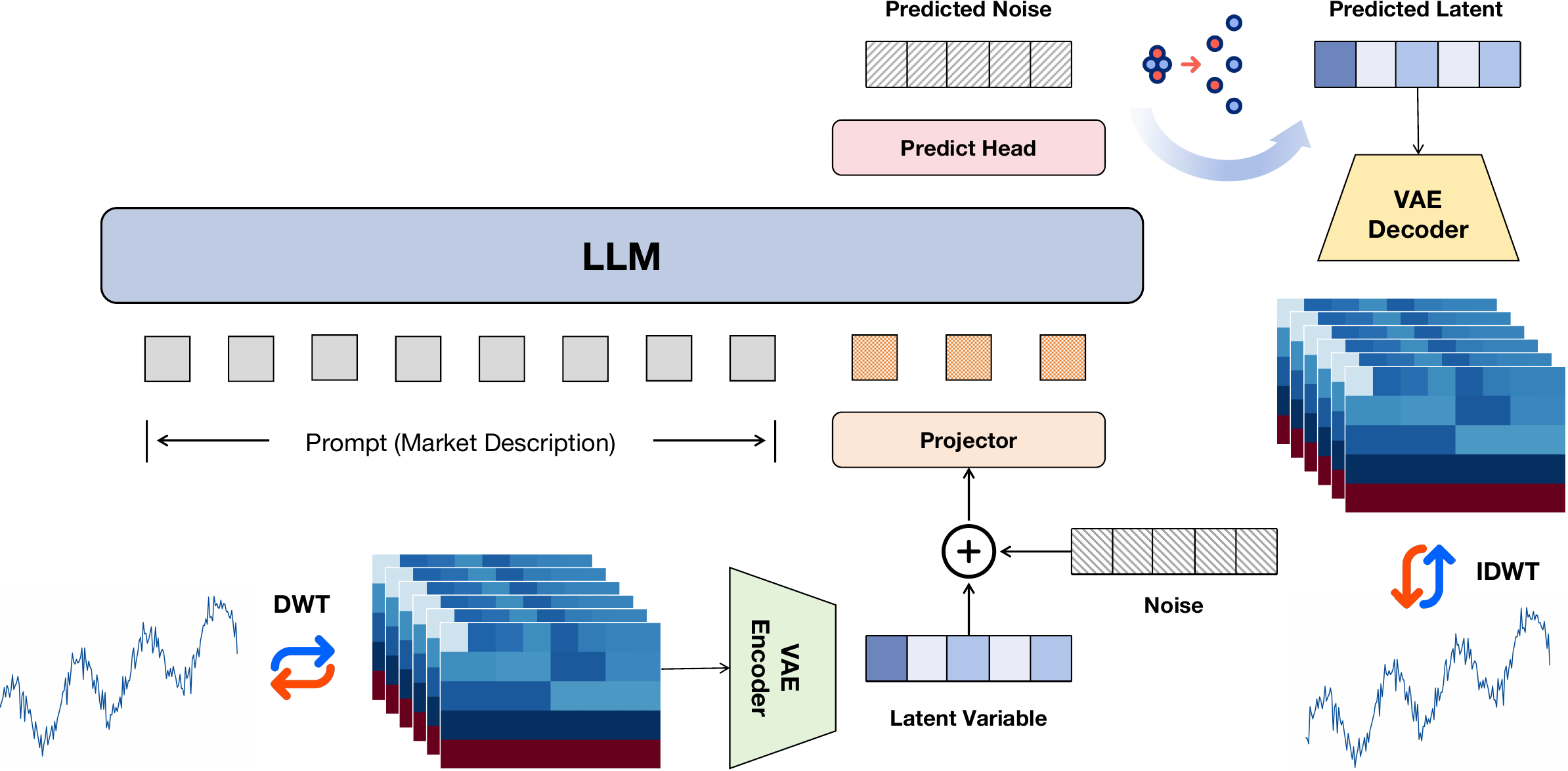}
\caption{Overview of the TF-CoDiT framework.}
\label{fig:2}
\end{figure}

\subsection{Backbone LLM}
The core architecture of TF-CoDiT is identical to FuseDiT~\cite{FuseDiT}, where a pre-trained LLM is used to model textual conditions and perform denoising. 
As shown in Figure~\ref{fig:2}, the inputs of this LLM consist of text embeddings followed by latent variables. 
At the attention layer $l$, the hidden states from the text $\textbf{h}_l^{text} \in \mathbb{R}^{N \times D}$ and the latent $\textbf{h}_l^{latent} \in \mathbb{R}^{M \times D}$ are concatenated together: $\textbf{h}_l = [\textbf{h}_l^{text}; \textbf{h}_l^{latent}] \in \mathbb{R}^{(N+M) \times D}$, and undergoes a shared self-attention with an asymmetric mask:
$$
\textbf{h}'_{l} = \text{Attention}(\text{LN}(\textbf{h}_l), \mathcal{M}) + \textbf{h}_l
$$
where $\text{LN}(\cdot)$ is layer normalization. $\mathcal{M}$ is a causal mask on text tokens, while a full mask on latent tokens:
$$
\mathcal{M}_{i,j} = 
\begin{cases} 
0 & \text{if } j \leq i \leq M \\
0 & \text{if } i > N \\
-\infty & \text{otherwise}
\end{cases}
$$
Subsequently, the two modalities diverge in their normalization and modulation: 
$$
\begin{array}{ll}
\textbf{h}_{l+1}^{text} &= \text{FFN}_{text}(\text{LN}(\textbf{h}_{l}^{text'})) + \textbf{h}_{l}^{text'}, \\
\textbf{h}_{l+1}^{latent} &= \text{FFN}_{latent}(\text{AdaLN}(\textbf{h}_{l}^{latent'}, t)) + \textbf{h}_{l}^{latent'}
\end{array}
$$
where an adaptive LN layer $\text{AdaLN}(\cdot)$ injects the diffusion timestep $t$ into the forward propagation.

\noindent \textbf{Training}. 
We define a forward diffusion process that adds noise $\epsilon$ to the latent variable $\textbf{z}_0$ over $T$ steps, resulting in $\textbf{z}_t$. The LLM is trained to predict the added noise via the following loss function:
$$
\mathcal{L}_{\text{diff}} = \mathbb{E}_{\textbf{z}_0, \epsilon \sim \mathcal{N}(0,\mathbf{I}), t} \left[ \| \epsilon - \epsilon_\theta(\textbf{z}_t, t, c) \|^2 \right]
$$
By freezing the LLM's pre-trained weights (i.e., only train the embedding and output layers), TF-CoDiT can capture the non-linear correlations between macro-economic narratives and the resulting price movements in the TF market.

\subsection{U-shape Variational Autoencoder}
TF-CoDiT necessitates a VAE model for: 1) bridging the gap between wavelet coefficients (a frequency variable) and the diffusion results (a latent variable), 2) modeling the elusive correlations among multiple time variables.

Before this work, U-Cast~\cite{U-Cast} introduced a \textit{Latent Query Attention} (LQA) mechanism that can effectively encode inter-channel dependencies of time series into the Transformer's hidden states. 
Inspired by U-Cast, we propose the U-shape VAE (U-VAE). 
It is a novel VAE architecture that embeds channels of the DWT spectrum into a dense vector.
Additionally, U-VAE shares learnable latent queries across the aligned layers of encoder and decoder. These queries vary in size with increasing layer depth, which effectively enables channel-wise encoding and recovery. 


\noindent \textbf{Encoder} $\mathcal{E}$.
Given the wavelet coefficients $\mathbf{W} \in \mathbb{R}^{C \times \Gamma \times T}$, each channel $\mathbf{W}_c$ is a structured time-frequency map, which aligns with ordinary images. 
To preserve local multi-scale features, we partition each channel into non-overlapping 2D patches of size $P_f \times P_t$. These patches are then projected into a sequence of time-and-frequency-specific vectors:
$$
\mathbf{h}_0^{(c)} = \text{Linear}(\text{Patchify}(\mathbf{W}_c)) + \mathbf{E}_{pos} \in \mathbb{R}^{N \times d_c}
$$
where $N = \frac{\Gamma}{P_f} \times \frac{T}{P_t}=N_f\times N_t$ is the number of pathces per channel, and $\mathbf{E}_{pos}$ are 2D positional embeddings that encode the time and frequency coordinates (see Appendix~\ref{sec:appendix}).
To capture the complex inter-channel dependencies while performing spatial compression, we follow U-Cast and build the U-VAE encoder by stacking $L$ Transformer layers. LQA is implemented at each layer $\ell$:
$$
\mathbf{Q} = \mathbf{W}_q^{(\ell)} \mathbf{Q}^{(\ell)},
\mathbf{K} = \mathbf{W}_k^{(\ell)} \mathbf{H}^{(\ell-1)},
\mathbf{V} = \mathbf{W}_v^{(\ell)} \mathbf{H}^{(\ell-1)}
$$
\[
\begin{aligned}
\mathbf{H}^{(\ell)} = \text{LN}(\text{MHA}(\mathbf{Q},\mathbf{K},\mathbf{V})) \in \mathbb{R}^{C_\ell \times d}
\end{aligned}
\]
where $\mathbf{H}^{(0)} = \text{Concat}(\mathbf{h}_{0}^{(1)}, \dots, \mathbf{h}_{0}^{(C)}) \in \mathbb{R}^{C\times d}$ are channel-wise embeddings, and $d=N\times d_c$. $\mathbf{Q}^{(\ell)} \in \mathbb{R}^{C_\ell \times d}$ is a learnable query, $C_\ell = C / r $, and $r\ge 1$ is a reduction ratio.
This hierarchical attention mechanism forces the model to distill global dependencies into an increasingly compact hidden space.
Consequently, the hidden state at the deepest layer, $\mathbf{H}^{(L)} \in \mathbb{R}^{1 \times d}(C_L=1)$, distills the multi-channel wavelet features into a compact representation. This state is then parameterized into the mean $\mu$ and variance $\sigma$ of the latent distribution, from which the latent variable $\textbf{z}_0$ is sampled:
$$
\textbf{z}_0 = \mu + \sigma \odot \epsilon, \quad \epsilon \sim \mathcal{N}(0, \mathbf{I})
$$
To align with the standard latent diffusion, we recover $\textbf{z}_0$ into a structured 3D spatial-temporal tensor $\mathcal{Z} \in \mathbb{R}^{N_f \times N_t \times d_c}$. 
The dimensions $N_f$ and $N_t$ scale with the frequency and temporal axes of the partitioned patches. 
This enables the diffusion generation on a structured grid that preserves the topological essence of the financial signals\footnote{Our diffusion model performs a similar 2D embedding.}.

\noindent \textbf{Decoder} $\mathcal{D}$.
The decoder inverts the compression process, transforming the sampled latent variable $\mathbf{z} \sim q(\mathbf{z}|\mathbf{W})$ back to the wavelet manifold while preserving multi-scale structures.
We first project $\mathbf{z}$ to match the encoder's deepest representation:
\[
\mathbf{U}^{(L)} = \text{Linear}_{\text{in}}(\mathbf{z}) \in \mathbb{R}^{C_L \times d}
\]
To restore the original channel dimension and reconstruct the wavelet coefficients, we employ an \textit{Up-sampling Latent Query Attention} that mirrors the encoder's compression hierarchy in reverse. At each decoding layer $\ell \in \{L, \dots, 1\}$:
$$
{\mathbf{Q}} = \mathbf{W}_q^{(\ell)} {\mathbf{Q}}^{(\ell)},
\tilde{\mathbf{K}} = \mathbf{W}_k^{(\ell)} \mathbf{U}^{(\ell)},
\tilde{\mathbf{V}} = \mathbf{W}_v^{(\ell)} \mathbf{U}^{(\ell)}
$$
\[
\mathbf{U}^{(\ell-1)} =\text{LN}(\text{MHA}(\tilde{\mathbf{Q}},\tilde{\mathbf{K}},\tilde{\mathbf{V}})) \in \mathbb{R}^{C_{\ell-1} \times d}
\]
where ${\mathbf{Q}}^{(\ell)}$ is drawn from the same layer of the encoder.
This architecture enables the decoder to selectively retrieve information from the compressed latent space while leveraging scale-aligned representations from the encoder for accurate reconstruction.
Finally, the reconstructed wavelet matrices are obtained through channel-wise projection and reshaping:
\[
\hat{\mathbf{W}} = \{ \text{Unpatchify}_{\Gamma  \times T}\left(\text{Linear}_{\text{out}}\left(\mathbf{U}^{(0)}_c\right)\right)\}_{c=1}^{C}
\]

\noindent \textbf{Learning Objective}.
U-VAE is trained by maximizing the evidence lower bound:
$$
\mathcal{L}_{\text{vae}} = \underbrace{\|\mathbf{W} - \hat{\mathbf{W}}\|_1}_{\text{Reconstruction term}} - \beta \cdot \underbrace{\text{KL}\left(q(\mathbf{z}|\mathbf{W}) \| p(\mathbf{z})\right)}_{\text{Regularization term}}
$$
Notably, we adopt an $L_1$ reconstruction loss (rather than the standard MSE) to promote sharpness in the reconstructed wavelet coefficients, ensuring that high-frequency "edges" in the time series are not smoothed out during the compression process, which is vital for high-fidelity synthesis in the TF-CoDiT framework.

\subsection{FinMAP}
The Financial Market Attribute Protocol (FinMAP) is not only an economic taxonomy that guides the tracing of market dynamics but also a structured language system, which facilitates writing prompt words that cover the essential conditions to promote treasury futures generation.

\begin{figure}[!t]
\centering
\begin{subfigure}[b]{0.46\linewidth}
\centering
\includegraphics[width=\linewidth]{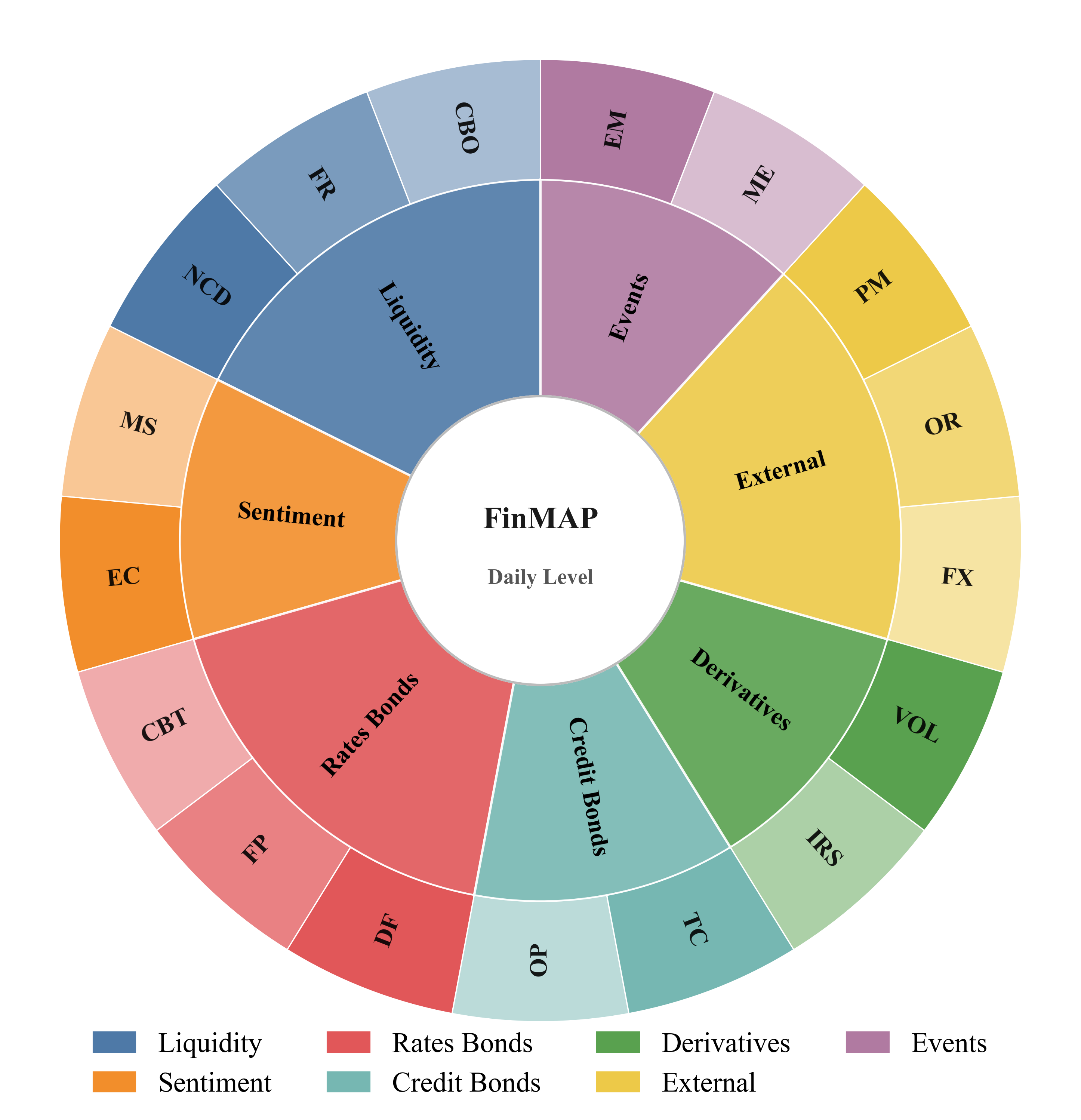}
\caption{Daily-level FinMAP}
\label{fig:sub1}
\end{subfigure}
\hfill
\begin{subfigure}[b]{0.52\linewidth}
\centering
\includegraphics[width=\linewidth]{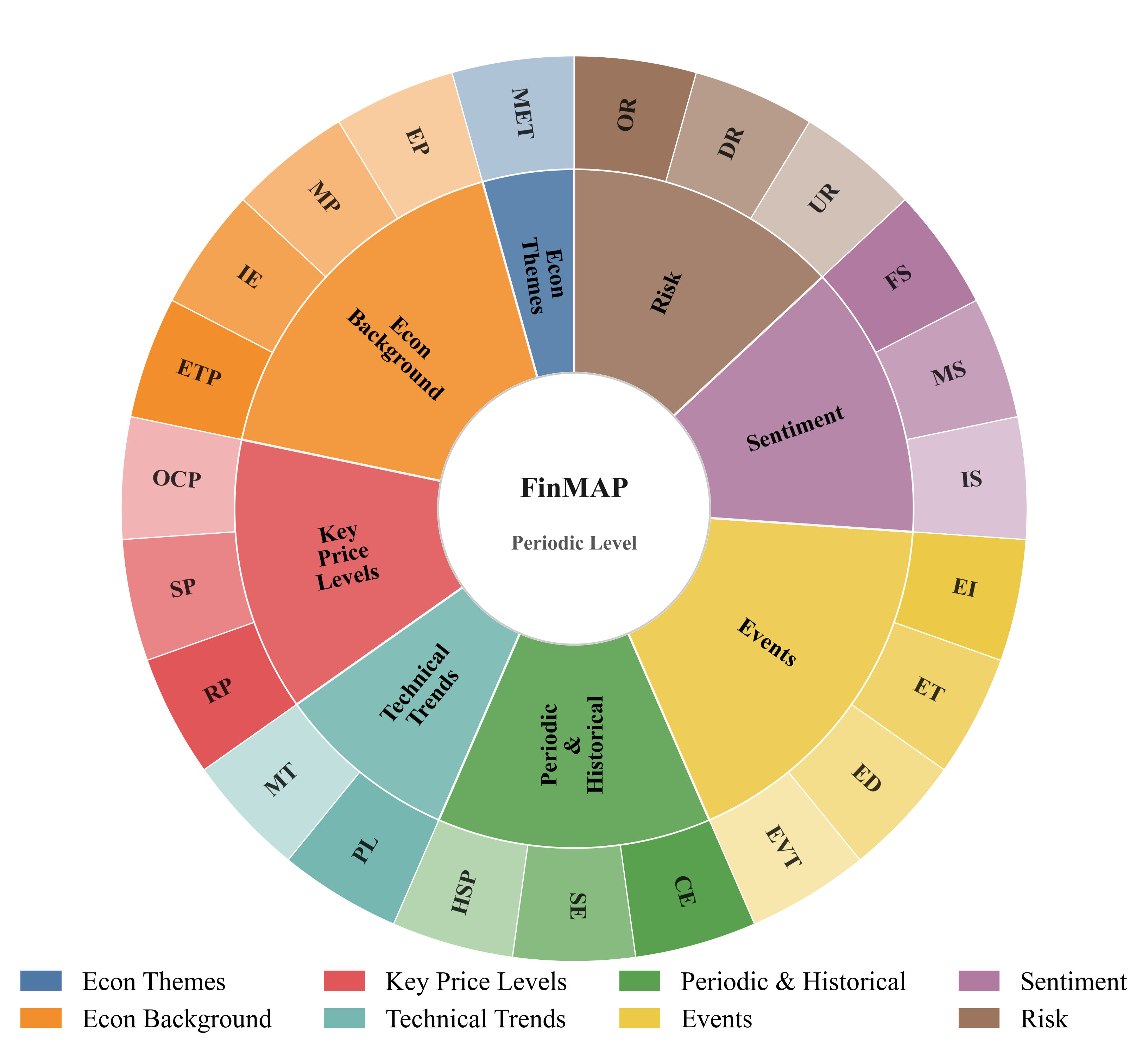}
\caption{Periodic-level FinMAP}
\label{fig:sub2}
\end{subfigure}
\caption{Architecture of FinMAP taxonomy.}
\vspace{-10pt}
\label{fig:fin-map}
\end{figure}

As illustrated in Figure~\ref{fig:fin-map}, the FinMAP's taxonomy is organized into a hierarchical architecture encompassing two distinct temporal resolutions: the daily snapshot and the periodic narrative.
\textbf{The daily snapshot} focuses on high-frequency liquidity and sentiment shifts that drive intra-day and day-to-day fluctuations. It categorizes market attributes into 7 primary categories, encompassing 17 types of economic factors. 
\textbf{The periodical narrative} captures long-term structural trends - spanning weeks, months, or quarters - that define the fundamental regime of the TF market. This level recognizes 23 distinct economic indicators reviewed from 8 perspectives. 
Based on this, we implement a temporal aggregation pipeline. Environmental data is first collected chronologically according to daily-level factors; subsequently, these daily snapshots are distilled and reformulated into periodic indicators. This process culminates in the generation of structured prompt anchors, which serve as the high-context conditioning inputs for TF-CoDiT, steering the model's generation toward financially grounded and temporally consistent market analysis.
(Appendix~\ref{sec:a2} provides more details).

\subsection{Inference}
The inference process of TF-CoDiT consists of three steps: 1) \textbf{Iterative Denoising}. Starting from Gaussian noise $\textbf{z}_T \sim \mathcal{N}(0, \mathbf{I})$, the LLM $\epsilon_\theta$ recovers the clean latent $\textbf{z}_0$ conditioned on a prompt $c$ following Fin-MAP. 
We employ the DPM-Solver~\cite{DPM-Solver} to solve the reverse-time stochastic differential equation:
$$
\textbf{z}_{t-1} = \Psi(\textbf{z}_t, \epsilon_\theta(\textbf{z}_t, t, c), t, \Delta t)
$$
where $\Psi$ is the solver transition.
2) \textbf{Latent Decoding}. The LLM-predicted latent $\textbf{z}_0$ is projected into the wavelet coefficient space via the pre-trained VAE decoder $\mathcal{D}$:
$$
\hat{\mathbf{W}} = \mathcal{D}(\textbf{z}_0), \quad \hat{\mathbf{W}} \in \mathbb{R}^{C \times L}
$$
3) \textbf{Signal Reconstruction}. We use the Inverse Discrete Wavelet Transform to map the coefficients back to the time domain and produce the 8-variable treasury futures sequence $\hat{X}$:
$$
\hat{X} = \text{IDWT}(\hat{\mathbf{W}}), \quad \hat{X} \in \mathbb{R}^{C \times T}
$$

\begin{table*}[ht]
\centering
\scriptsize
\caption{
Evaluation results of time series synthesis for contracts TS, TF, T, and TL.
Best results are in \textbf{bold}, second best are \underline{underlined}.
Outlier results are in \textcolor{red}{red}. The maximum errors of each model are in \textcolor{blue}{blue}.
}
\label{tab:1}
\begin{tabular}{c|c|ccc|ccc|ccc|ccc}
\toprule
\multirow{2.5}{*}{\textbf{Models}} & \multirow{2.5}{*}{\textbf{Metric}} & \multicolumn{3}{c|}{TS} & \multicolumn{3}{c|}{TF} & \multicolumn{3}{c|}{T} & \multicolumn{3}{c}{TL} \\ 
\cmidrule{3-5}\cmidrule{6-8}\cmidrule{9-11}\cmidrule{12-14}
 & & 8 & 32 & 64 & 8 & 32 & 64 & 8 & 32 & 64 & 8 & 32 & 64 \\
\midrule
\multirow{2}{*}{TimeGAN} & MSE & {0.408} & {0.423} & 0.465 & 0.455 & 0.422 & 0.437 & {0.413} & {0.430} & 0.458 & 0.450 & \textcolor{blue}{0.480} & 0.460 \\
 & MAE & {0.434} & {0.434} & {0.483} & 0.481 & 0.412 & 0.431 & 0.446 & {0.430} & {0.479} & 0.514 & 0.527 & \textcolor{blue}{0.537} \\
\midrule
\multirow{2}{*}{WGAN} & MSE & {0.429} & {0.472} & 0.488 & 0.403 & 0.457 & {0.478} & {0.451} & \textcolor{blue}{0.480} & 0.400 & 0.406 & 0.448 & 0.452 \\
 & MAE & {0.251} & {0.413} & 0.484 & 0.439 & 0.467 & 0.433 & {0.455} & \underline{0.415} & 0.491 & 0.533 & 0.505 & \textcolor{blue}{0.549} \\
\midrule
\multirow{2}{*}{QuatGAN} & MSE & {0.525} & {0.557} & 0.537 & \textcolor{blue}{0.570} & 0.548 & 0.500 & {0.425} & {0.464} & 0.459 & 0.487 & 0.509 & 0.560 \\
 & MAE & {0.458} & {0.452} & 0.467 & 0.463 & 0.466 & 0.463 & {0.561} & {0.552} & 0.592 & 0.595 & 0.514 & \textcolor{blue}{0.627} \\
\midrule
\multirow{2}{*}{GBMDiff} & MSE & {0.388} & {0.664} & 0.614 & 0.794 & \textcolor{red}{1.433} & \textcolor{red}{1.295} & {0.390} & {0.863} & 0.920 & 0.936 & \textcolor{red}{1.610} & \textcolor{red}{1.376} \\
 & MAE & {0.435} & {0.601} & {0.625} & 0.903 & \textcolor{red}{1.169} & \textcolor{red}{1.041} & {0.443} & {0.797} & \textcolor{red}{1.139} & 0.931 & \textcolor{red}{1.847} & \textcolor{red}{1.144} \\
\midrule
\multirow{2}{*}{FinDDPM} & MSE & {0.456} & {0.422} & {0.490} & {0.425} & 0.491 & 0.400 & 0.433 & 0.442 & \textcolor{blue}{0.669} & 0.428 & 0.439 & 0.461 \\
 & MAE & {0.370} & {0.394} & {0.397} & {0.421} & 0.405 & \textcolor{blue}{0.538} & 0.415 & 0.431 & 0.479 & 0.465 & 0.466 & 0.475 \\
\midrule
\multirow{2}{*}{TimeDiT$^{*}$} & MSE & {0.404} & {0.427} & 0.464 & 0.441 & {0.311} & {0.329} & 0.301 & 0.366 & 0.277 & {0.437} & \textcolor{blue}{0.522} & 0.505 \\
 & MAE & {0.277} & {0.323} & \underline{0.293} & {0.335} & 0.316 & 0.368 & 0.296 & 0.343 & 0.320 & 0.353 & 0.463 & \textcolor{blue}{0.488} \\
\midrule
\multirow{2}{*}{T2S$^{*}$} & MSE & \textbf{0.244} & \textbf{0.293} & \textbf{0.283} & {0.289} & 0.241 & 0.383 & \textbf{0.262} & 0.297 & \underline{0.275} & \underline{0.404} & \textcolor{blue}{0.553} & \underline{0.424} \\
 & MAE & \textbf{0.175} & {0.370} & {0.376} & {0.369} & 0.380 & 0.380 & \underline{0.380} & 0.373 & 0.423 & 0.392 & \underline{0.346} & \textcolor{blue}{0.427} \\
\midrule
\multirow{2}{*}{TimeLLM$^{*}$} & MSE & {0.429} & \underline{0.306} & 0.440 & 0.310 & 0.447 & 0.313 & {0.430} & {0.305} & 0.451 & 0.535 & \textcolor{blue}{0.663} & 0.525 \\
 & MAE & {0.356} & \underline{0.322} & \textcolor{blue}{0.390} & \textbf{0.225} & \underline{0.291} & \underline{0.300} & 0.333 & 0.342 & \textbf{0.269} & \underline{0.328} & \textbf{0.339} & \underline{0.361} \\
\midrule
\multirow{2}{*}{CALF$^{*}$} & MSE & {0.370} & {0.394} & {0.397} & {0.421} & \textcolor{blue}{0.605} & 0.538 & 0.415 & 0.431 & 0.479 & 0.465 & 0.466 & 0.475 \\
 & MAE & {0.404} & {0.427} & 0.464 & 0.441 & {0.411} & {0.429} & 0.501 & {0.466} & 0.677 & 0.537 & \textcolor{blue}{0.722} & 0.605 \\
\midrule
\multirow{2}{*}{TF-CoDiT$^{*}$} & MSE & \underline{0.277} & {0.323} & \underline{0.293} & \textbf{0.235} & \textbf{0.216} & \textbf{0.268} & \underline{0.296} & \textbf{0.243} & \textbf{0.220} & \textbf{0.353} & \textcolor{blue}{\textbf{0.433}} & \textbf{0.418} \\
 & MAE & \underline{0.234} & \textbf{0.273} & \textbf{0.283} & \underline{0.275} & \textbf{0.241} & \textbf{0.283} & \textbf{0.242} & \textbf{0.279} & \underline{0.279} & \textbf{0.301} & \textcolor{blue}{0.384} & \textbf{0.324} \\
\bottomrule
\end{tabular}
\end{table*}

\section{Experiment}

\subsection{Datasets}
We constructed a comprehensive dataset comprising treasury futures contracts for TS (2-year), TF (5-year), T (10-year), and TL (30-year), covering the period from March 20, 2015, to December 31, 2025. This dataset consists of four distinct subsets with the daily observations: 1,789 for TS, 2,995 for TF, 2,625 for T, and 655 for TL.
To facilitate the training and evaluation of our model, we created high-quality prompt-time series pairs by aggregating daily market reviews from the Wind Terminal\footnote{\url{https://www.wind.com.cn/}} and our internal proprietary databases. 
Additionally, an automatic labeling pipeline is used to extract these unstructured narrative reviews into our structured FinMAP protocol. 
Appendix~\ref{sec:a3} details specifications regarding the labeling heuristics and data cleaning procedures.

\subsection{Experimental Settings}
In experiments, we evaluate the model's capability to generate TF time series across multiple temporal horizons, ranging from one business week to one fiscal quarter. 
To facilitate the implementation, we run the model to generate sequences of lengths $L \in \{8, 32, 64, 128\}$ and report the averaged \textbf{MSE} and \textbf{MAE} scores across opening/highest/lowest/closing prices (OHLC). The test set comprises time series data spanning the last 200 days in our dataset. Appendix~\ref{sec:data} provides more details.

The prompt construction follows a hierarchical temporal aggregation strategy: for the 8-day generation task, the prompt absorbs the information from 8 consecutive daily FinMAP attributes. For extended horizons, such as the 32-day or 64-day tasks, the prompts are constructed by recursively aggregating lower-level periodic FinMAP records (e.g., a 16-day prompt aggregates two 8-day periodic situational descriptions). 
This hierarchical prompt engineering ensures that the model maintains a consistent "macro-narrative" while accounting for long-term structural shifts in the market regime.

\textbf{Comparison models}.
We compared TF-CoDiT with three types of models: 
(1) \textbf{Time-series GANs}: TimeGAN~\cite{Time-GAN}, WGAN~\cite{WGAN}, and QuatGAN~\cite{QuantGAN}. 
(2) \textbf{Diffusion Models}: GBMDiff~\cite{DM-based} and FinDDPM~\cite{DDPM-based}, TimeDiT~\cite{TimeDiT}, T2S~\cite{T2S}. 
(3) \textbf{Time-series Foundation Models}: TimeLLM~\cite{Time-LLM}, CALF~\cite{CALF}.
Appendix~\ref{sec:a4} details our training strategy and implementation of these models.


\begin{table*}[ht]
\centering
\scriptsize
\caption{Evaluation results of 128-day time series synthesis.}
\label{tab:2}
\begin{tabular}{cccccccccccc}
\toprule
\textbf{Contract} & \textbf{Metric} & TimeGAN & WGAN & QuatGAN & GBMDiff & FinDDPM & TimeDiT$^{*}$ & T2S$^{*}$ & TimeLLM$^{*}$ & CALF$^{*}$ & TF-CoDiT$^{*}$ \\ 
\midrule
\multirow{2}{*}{TS} & MSE & {0.475} & {0.470} & {0.476} & {0.699} & 0.480 & 0.480 & 0.480 & 0.473 & \underline{0.423} & \textbf{0.402} \\
 & MAE & {0.469} & {0.496} & \underline{0.440} & 0.710 & 0.447 & 0.513 & {0.430} & {0.505} & 0.451 & \textbf{0.405} \\
\midrule
\multirow{2}{*}{TF} & MSE & \textcolor{blue}{0.627} & \textcolor{blue}{0.543} & \textcolor{blue}{0.681} & \textcolor{blue}{0.750} & \textcolor{blue}{0.560} & \textcolor{blue}{0.611} & \textcolor{blue}{0.494} & \textcolor{blue}{0.569} & \textbf{0.425} & \underline{\textcolor{blue}{0.447}}  \\
 & MAE & \textcolor{blue}{0.582} & \textcolor{blue}{0.618} & \textcolor{blue}{0.600} & \textcolor{blue}{0.733} & \textcolor{blue}{0.694} & \textcolor{blue}{0.727} & \textcolor{blue}{0.577} & \textcolor{blue}{0.615} & \textbf{0.439} & \underline{0.448} \\
\midrule
\multirow{2}{*}{T} & MSE & {0.525} & {0.434} & 0.572 & 0.550 & \underline{0.400} & {0.417} & 0.426 & 0.476 & 0.417 & \textbf{0.391} \\
 & MAE & {0.474} & {0.423} & {0.568} & {0.546} & 0.399 & 0.426 & \textbf{0.314} & 0.352 & \underline{0.444} & 0.372 \\
\midrule
\multirow{2}{*}{TL} & MSE & {0.560} & 0.509 & {0.563} & {0.601} & \underline{0.463} & 0.508 & 0.469 & {0.503} & \textcolor{blue}{0.498} & \textbf{0.418} \\
 & MAE & {0.579} & {0.568} & {0.578} & {0.583} & 0.476 & \underline{0.475} & 0.481 & 0.477 & \textcolor{blue}{0.502} & \textbf{\textcolor{blue}{0.453}} \\
\bottomrule
\end{tabular}
\end{table*}


\begin{table*}[ht]
\centering
\scriptsize
\caption{
Evaluation results of conditioned (Cond.) generation versus unconditional generation of TF-CoDiT.
}
\label{tab:3}
\begin{tabular}{c|c|ccc|ccc|ccc|ccc}
\toprule
\multirow{2.5}{*}{\textbf{Cond.}} & \multirow{2.5}{*}{\textbf{Metric}} & \multicolumn{3}{c|}{TS} & \multicolumn{3}{c|}{TF} & \multicolumn{3}{c|}{T} & \multicolumn{3}{c}{TL} \\ 
\cmidrule{3-5}\cmidrule{6-8}\cmidrule{9-11}\cmidrule{12-14}
 & & 8 & 32 & 64 & 8 & 32 & 64 & 8 & 32 & 64 & 8 & 32 & 64 \\
\midrule
\multirow{2}{*}{\ding{51}} & MSE & {0.277} & {0.323} & {0.293} & {0.235} & {0.216} & {0.268} & {0.296} & {0.243} & {0.220} & {0.353} & {0.463} & {0.418} \\
 & MAE & {0.234} & {0.273} & {0.238} & {0.275} & {0.241} & {0.283} & {0.242} & {0.279} & {0.279} & {0.301} & 0.384 & {0.324} \\
\midrule
\multirow{2}{*}{\ding{55}} & MSE & 0.448 & 0.415 & 0.482 & 0.431 & 0.485 & 0.407 & 0.426 & 0.437 & 0.462 & 0.422 & 0.444 & 0.458 \\
 & MAE & 0.376 & 0.389 & 0.402 & 0.418 & 0.411 & 0.529 & 0.422 & 0.436 & 0.473 & 0.459 & 0.471 & 0.469 \\
\bottomrule
\end{tabular}
\end{table*}

\begin{table*}[!ht]
\centering
\scriptsize
\caption{
Evaluation results of TF-CoDiT using FinMAP, partial FinMAP, and without using FinMAP.
}
\label{tab:4}
\begin{tabular}{c|c|c|ccc|ccc|ccc|ccc}
\toprule
\multicolumn{2}{c|}{\textbf{FinMAP}} & \multirow{2.5}{*}{\textbf{Metric}} & \multicolumn{3}{c|}{TS} & \multicolumn{3}{c|}{TF} & \multicolumn{3}{c|}{T} & \multicolumn{3}{c}{TL} \\ 
\cmidrule{1-2}\cmidrule{4-6}\cmidrule{7-9}\cmidrule{10-12}\cmidrule{13-15}
Daily & Period & & 8 & 32 & 64 & 8 & 32 & 64 & 8 & 32 & 64 & 8 & 32 & 64 \\
\midrule
\multirow{2}{*}{\ding{51}} & \multirow{2}{*}{\ding{51}} & MSE & {0.277} & {0.323} & {0.293} & {0.235} & {0.216} & {0.268} & {0.296} & {0.243} & {0.220} & {0.353} & {0.463} & {0.418} \\
 & & MAE & {0.234} & {0.273} & {0.238} & {0.275} & {0.241} & {0.283} & {0.242} & {0.279} & {0.279} & {0.301} & 0.384 & {0.324} \\
\midrule
\multirow{2}{*}{\ding{55}} & \multirow{2}{*}{\ding{51}} & MSE & 0.391 & 0.419 & 0.379  & 0.331 & 0.491 & 0.438 & 0.392 & 0.447 & {0.383} & {0.353} & 0.473 & 0.462 \\
 &  & MAE & {0.362} & {0.392} & 0.376 & {0.397} & {0.398} & {0.418} & 0.387 & 0.439 & {0.383} & {0.372} & 0.388 & 0.403 \\
\midrule
\multirow{2}{*}{\ding{51}} & \multirow{2}{*}{\ding{55}} & MSE & 0.402 & 0.417 & 0.398 & 0.347 & 0.483 & 0.447 & 0.421 & 0.461 & {0.461} & {0.411} & 0.417 & 0.421 \\
 &  & MAE & 0.388 & 0.420 & 0.385 & 0.327 & 0.476 & 0.429 & 0.416 & 0.418 & {0.429} &  0.408 & 0.415 & 0.423 \\
\midrule
\multirow{2}{*}{\ding{55}} & \multirow{2}{*}{\ding{55}} & MSE & 0.448 & 0.415 & 0.482 & 0.431 & 0.485 & 0.407 & 0.426 & 0.437 & 0.462 & 0.422 & 0.444 & 0.458 \\
 & & MAE & 0.376 & 0.389 & 0.402 & 0.418 & 0.411 & 0.529 & 0.422 & 0.436 & 0.473 & 0.459 & 0.471 & 0.469 \\
\bottomrule
\end{tabular}
\vspace{-5pt}
\end{table*}

\subsection{Evaluation Results}

Table~\ref{tab:1} reports the TF time-series synthesis results across temporal horizons ranging from 8 to 64 days. 
A notable observation is that the specific contract type does not significantly impact model performance, except for TL due to its exclusively limited volume (see Table~\ref{tab:data}).
GAN-based models (e.g., TimeGAN) exhibit higher error rates, with MSE values often exceeding 0.45 and MAE around 0.5. In contrast, diffusion models and foundation models perform much better; for instance, the average 32-day MSE of FinDDPM is 0.42. 
Moreover, conditional generative models (marked with $^{*}$) exhibit significant superiority across the contract TS, TF, and T.
TF-CoDiT outperforms autoregressive models TimeLLM and CALF by addressing the prevalent error accumulation. It also surpasses the previous best model, T2S, with an average improvement of 13.4\% in MSE and 12.8\% in MAE. Particularly, on the 32-day horizon, TF-CoDiT achieves an average MSE of 0.31 and an MAE of 0.29, whereas T2S achieves 0.35 and 0.37, respectively.

In Table~\ref{tab:2}, we further compare the models' performance in generating ultra-long time series of 128 days. While most of the compared models experienced significant performance degradation - the average MSE of T2S climbing to 0.47 - TF-CoDiT exhibited robustness with a leading MSE of 0.41 and MAE of 0.42. 

\textbf{Ablation Studies} are first conducted to investigate the necessity of conditioned generation. 
As reported in Table~\ref{tab:3}, removing the semantic prompts degrades the performance of TF-CoDiT to a level comparable to the unconditional FinDDPM, achieving an average MSE of 0.44 and MAE of 0.43 in 32-day generation tasks. These errors increase 0.14 (0.3  $\to$ 0.44) and 0.16 (0.27 $\to$ 0.43) compared to the standard TF-CoDiT. 
Given that text conditions are required, we further evaluate the functions of our FinMAP. 
We drop the daily FinMAP by asking the labeling agent to generate free-form prompts directly from raw market reviews, bypassing our structured taxonomy. To drop the periodic FinMAP, the model generates a macro-narrative that summarizes all daily reviews within the span, bypassing our recursive aggregation logic. 
Table~\ref{tab:4} indicates that the daily-level FinMAP is a primary driver of performance. 
With the daily-level FinMAP, the average MSE of unconditional generations declines 0.01 (row 3 vs. row 4); it also improves the generations with solely periodic FinMAP by reducing 0.11 average MSE (row 1 vs. row 2).

\begin{figure}[!t]
\centering
\begin{subfigure}[b]{0.95\linewidth}
\centering
\includegraphics[width=\linewidth]{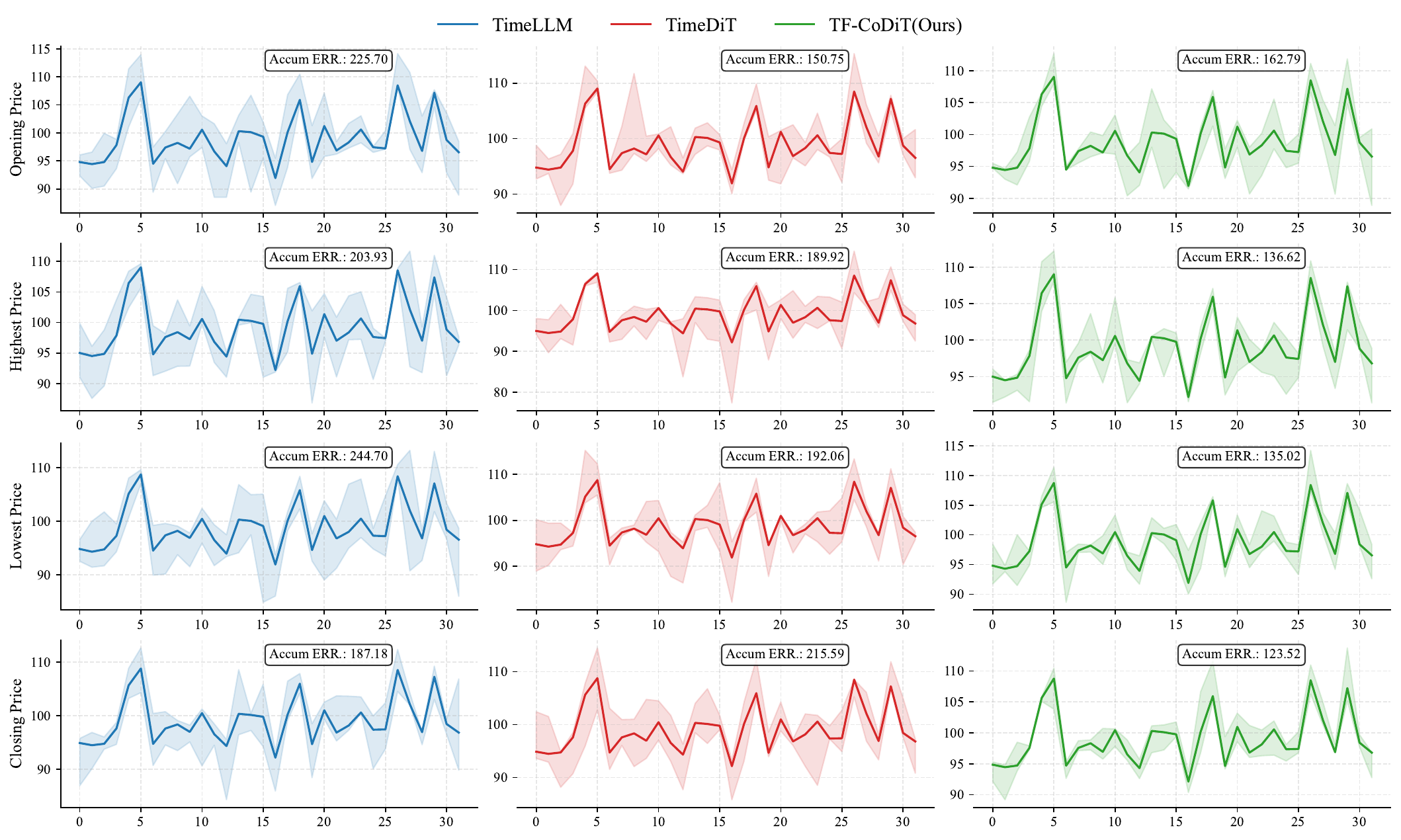}
\caption{Error band of 32-day generation.}
\label{fig:4-1}
\end{subfigure}
\vspace{1em} 
\begin{subfigure}[b]{0.95\linewidth}
\centering
\includegraphics[width=\linewidth]{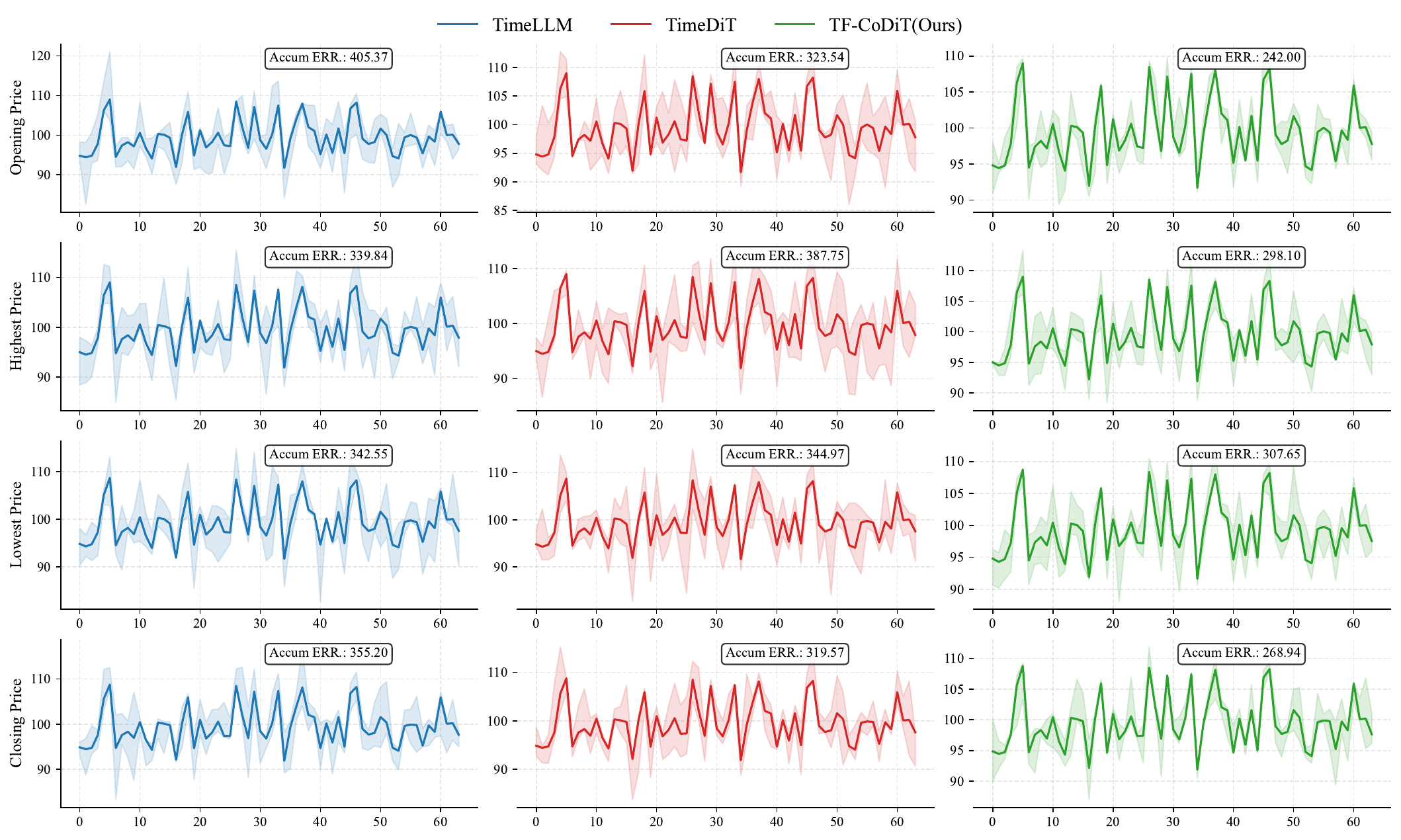}
\caption{Error band of 64-day generation.}
\label{fig:4-2}
\end{subfigure}
\vspace{-15pt}
\caption{Error band analyses in the contract-T.}
\vspace{-10pt}
\label{fig:4}
\end{figure}

\textbf{Case Study}.
In Figure~\ref{fig:4}, we compare TimeLLM, TimeDiT, and TF-CoDiT by generating 20 trajectories for the T-contract across 32-day and 64-day horizons. We analyze the resulting error bands relative to the ground-truths.
TF-CoDiT achieves the lowest cumulative reconstruction error, outperforming TimeDiT by 190.37 in the 32-day setting. Notably, this performance gap more than doubles at the 64-day horizon. This non-linear scaling demonstrates the robustness of TF-CoDiT in suppressing the variance explosion typically observed in long-term financial generation.

\section{Discussions}

\begin{figure}[!t]
\centering
\includegraphics[width=0.95\linewidth]{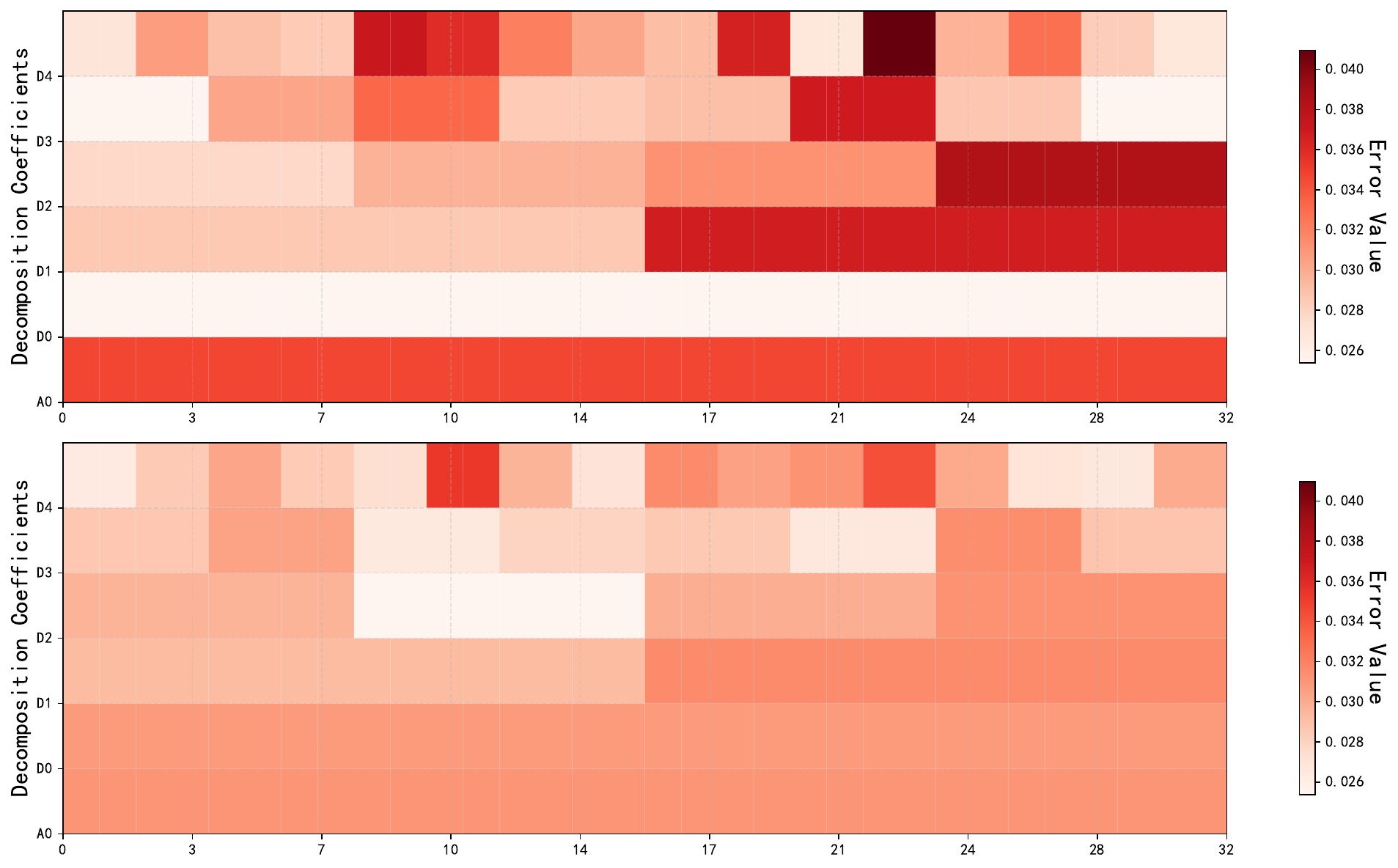}
\caption{
Element-wise reconstruction errors achieved by U-VAE with MSE loss (upper) and L1 loss (lower).
}
\label{fig:heat}
\end{figure}

\textbf{MSE Loss vs. L1 Loss in U-VAE}. We evaluate the impact of the reconstruction loss on the U-VAE by comparing the widely adopted MSE loss against the L1 loss. Figure~\ref{fig:heat} visualizes the element-wise reconstruction errors for the DWT matrix of opening price, with the 32-day T-contract. 
L1 loss significantly reduces reconstruction errors, achieving an 1.24\% (0.0304 $\to$ 0.0308) reduction in approximation and 10.94\% (0.0275 $\to$ 0.0309) in detail ($D_0$) coefficients compared to MSE. 
This improvement stems from the \textbf{heavy-tailed} distribution of treasury futures. MSE over-penalizes outliers, resulting in \textit{blurred} reconstructions that average out sharp transitions. In contrast, by maintaining a constant gradient, L1 preserves high-frequency \textit{detail} coefficients essential for capturing intraday volatility and structural breakouts.

\begin{figure}[!t]
\centering
\includegraphics[width=0.95\linewidth]{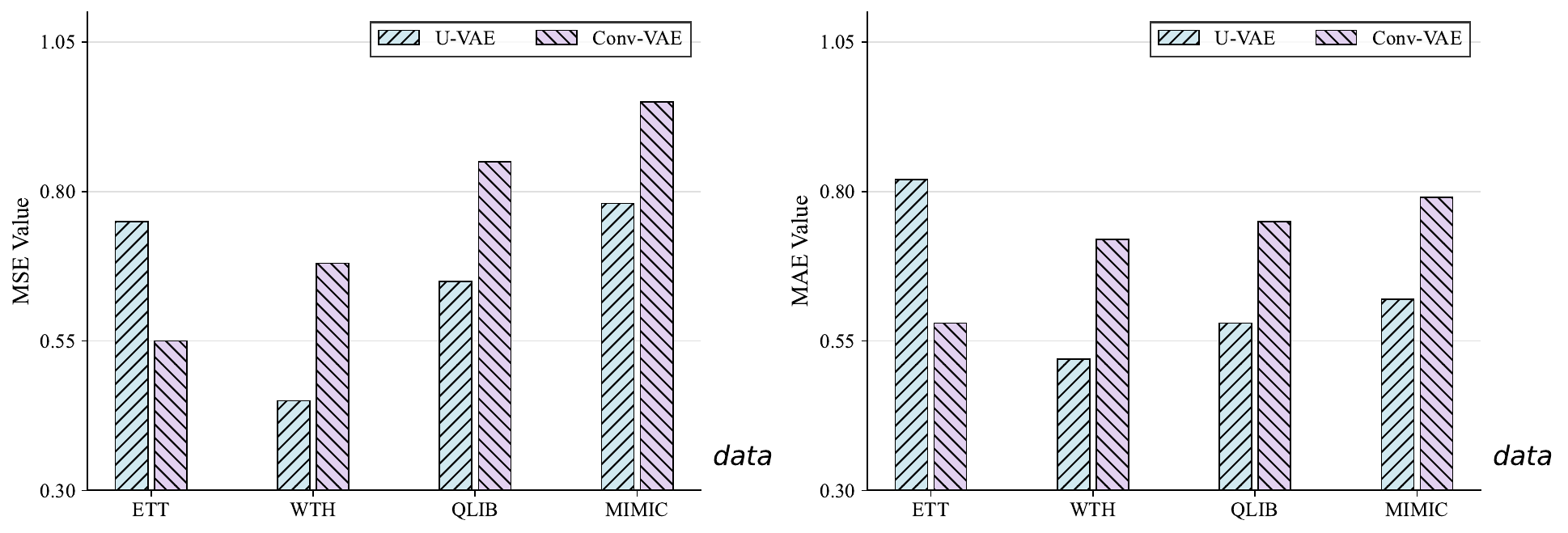}
\caption{Reconstruction errors over DWT matrix.}
\vspace{-10pt}
\label{fig:error}
\end{figure}

\textbf{Convolutional VAE vs. U-VAE}.
To test the effectiveness of our proposed channel-wise encoding technology, we compare hierarchical LQA-based U-VAE with the Conv-VAE widely used for image encoding (which consists of four 2D convolutional layers in both the encoder and decoder, with channels $c \in [128, 256, 512, 512]$ across the layers). Their reconstruction losses across four domain-specific multivariate time series datasets, ETT~\cite{ETT}, WTH~\cite{WHT}, QLIB~\cite{Qlib}, and MIMIC~\cite{MIMIC}, are compared in Figure~\ref{fig:error}. On three of these datasets, our model achieved better results. This comparison empirically proves the importance of modeling cross-channel correlations.

\section{Conclusion}
\label{sec:7}

This work introduces TF-CoDiT, a novel LLM-DiT framework for conditioned time-series generation (TSG) of treasury futures. 
We establish a robust link between linguistic narratives of economics and time-continuous prices modeling by: 1) a U-shaped VAE that encodes cross-channel correlations via latent query attention; and 2) the FinMAP protocol, which provides a criterion that standardizes contextual prompts. Experimental results across real contracts demonstrate that TF-CoDiT significantly outperforms existing state-of-the-art TSG models, achieving improved fidelity of generation results.


\newpage
\textbf{Limitations}
While TF-CoDiT exhibits superiority over existing time series generation models, it possesses several limitations that warrant further investigation:
\begin{itemize}
\item \textbf{Domain Knowledge Integration}. As the core architecture of TF-CoDiT is adapted from a text-to-image model, the pre-trained parameters lack intrinsic alignment with complex financial and economic theories. This gap necessitates extensive fine-tuning and can hinder the model’s ability to generalize to "black-swan"-like events or rare market regimes not heavily observed in the training distribution.
\item \textbf{Unimodal Conditioning Constraints}. TF-CoDiT relies exclusively on textual narratives to steer market dynamics. While this interface is human-friendly, textual prompts are inherently discrete and information-sparse, compared to continuous cross-asset signals. Incorporating multi-modal conditioners—such as synchronous time series of commodity prices (e.g., gold and oil) or foreign exchange rates remains a promising but unexplored avenue for further enhancing synthesis fidelity.
\item \textbf{Evaluation of FinMAP}. While the efficacy of the FinMAP protocol is empirically validated through our ablation studies, its evaluation remains largely intrinsic. Currently, there is a lack of an external, standardized benchmark to quantitatively assess the semantic quality and coverage of such financial taxonomies. Establishing a rigorous, objective benchmark for market-description protocols is a vital task left for our future work.
\end{itemize}

\bibliography{custom}

\appendix
\section{Appendix}
\label{sec:appendix}

\subsection{Treasury Futures}
\label{sec:a0}
Treasury futures (TF) are standardized exchange-traded contracts to buy or sell a government bond (the underlying asset) at a predetermined price on a specified future date. They are a primary type of interest rate future, originating in the 1970s to help investors manage the risk of fluctuating interest rates.
As a pivotal instrument for interest rate risk management, their pricing dynamics are fundamentally derived from the present value of the underlying bond's cash flows, making them highly sensitive to yield curve fluctuations.
Contracts are categorized by the maturity of the notional deliverable bond. Table~\ref{tab:tf} outlines the key structures in China and the United States - the two largest markets in the world.

This work focuses on treasury futures traded on the China Financial Futures Exchange (CFFEX). Additionally, the dynamics of the TF market are traced by eight key daily variables:
\begin{itemize}
\item \textbf{Opening price (O)}. The price at which the first transaction of the trading day occurs, indicating the initial market sentiment.
\item \textbf{Highest price (H)}. The maximum transaction price recorded during the trading session, reflecting the upper boundary of intraday price volatility.
\item \textbf{Lowest price (L)}. The minimum transaction price recorded during the trading session, reflecting the lower boundary of intraday price volatility.
\item \textbf{Close price (C)}. The price of the last transaction before the market closes, widely used as a daily benchmark.
\item \textbf{Settlement price}. The official price calculated by the exchange at the close, typically as a volume-weighted average, which serves as the daily mark-to-market benchmark for margin calculations.
\item \textbf{Trading value}. The total monetary value of all contracts traded during the day (measured in millions of CNY in the sample), representing the overall market activity scale.
\item \textbf{Trading volume}. The total number of contracts traded during the day (labeled as "stock" in the sample but representing the standardized contract unit), reflecting market liquidity and participation intensity.
\item \textbf{Open interest}. The total number of outstanding contracts held by market participants at the end of the trading day, indicating the market's depth and the persistence of capital commitments.
\end{itemize}

Together, these variables provide a comprehensive view of price action, liquidity, and market structure for the CFFEX treasury futures market. Particularly, the \textbf{opening price}, \textbf{highest price}, \textbf{lowest price}, and \textbf{close price} (OHLC) are pivotal for valuation, while the remaining variables collectively gauge trading activity and market sentiment.

\begin{table*}[!t]
\centering
\small
\caption{Overview of treasury futures contracts.}
\begin{tabular}{c|c|l}
\toprule
\textbf{Contract} & \textbf{Maturity} & \makecell{\centering \textbf{Features}} \\
\midrule
\multicolumn{3}{c}{\textbf{China (Traded on China Financial Futures Exchange)}} \\
\midrule
TS	& 2-Year & Tracks short-term rates \\
TF	& 5-Year & Tracks medium-term rates \\
T	& 10-Year & Tracks benchmark long-term rates \\
TL	& 30-Year & Tracks bltra-long term rates \\
\midrule
\multicolumn{3}{c}{\textbf{United States (Traded on CME Group)}} \\
\midrule
T-Note Futures & 2-year, 5-year, 10-year & Tracks medium to long-term yields \\
T-Bond Futures & 30-year & Tracks long-term interest rate expectations \\
\bottomrule
\end{tabular}
\label{tab:tf}
\end{table*}

\subsection{Time-Frequency Position Embedding}
\label{sec:a1}
The structured map $M$, resulting from our Multi-scale Temporal Alignment (MTA), presents a grid-like topology. We partition this map into non-overlapping patches of size $P_l \times P_t$. Each patch acts as a single discrete token in the Transformer, where $(I, J)$ denotes its grid coordinates in the frequency-temporal domain: $I \in \{0,1, \dots, \frac{J+1}{P_f}-1\}$ and $J \in \{0,1, \dots, \frac{R}{P_t}-1\}$ ($R=\frac{1}{T}$ in our implementation is the scaled time steps after DWT).
To ground these tokens in the global time-frequency space, we implement a composite embedding strategy. First, the content of each patch is aggregated via a linear projection layer. This layer effectively captures the local relative positioning within the patch. Then, we assign a 2D axial embedding based on the patch's grid indices:
$$
\begin{aligned}
\mathbf{E}_{I,J} =& \text{Linear}(\text{flatten}(\text{Patch}_{I,J})) + \mathbf{PE}_{freq}(I) \\
& + \mathbf{PE}_{time}(J)
\end{aligned}
$$
where $\mathbf{PE}_{freq}(I) \in \mathbb{R}^D$ is the Frequency-axis Embedding. It assigns a unique signature to each row of the grid. This allows the self-attention mechanism to recognize that tokens in the same row share the same spectral scale (e.g., long-term trends vs. short-term noise), facilitating intra-scale (horizontal) dependency modeling.
$\mathbf{PE}_{time}(J) \in \mathbb{R}^D$ is the Time-axis Embedding. 
It assigns a unique signature to each column. This enables the model to identify tokens across different frequency bands that occur at the same logical timestamp, thereby enhancing inter-scale (vertical) synchronization.

By assigning embeddings at the patch level rather than the element level, we allow the model to capture high-level structural correlations while significantly reducing the computational complexity of the self-attention mechanism. This coordinate system ensures that the model can effectively learn complex cross-scale correlations - such as how a macro-economic shift in the low-frequency trend layer triggers specific volatility patterns in high-frequency detail layers - while maintaining strict temporal alignment. This spatial awareness is crucial for synthesizing 8-variable treasury futures that are both statistically plausible and economically consistent.

\subsection{Financial Market Attribute Protocol}
\label{sec:a2}

FinMAP is a description system that facilitates the development of prompt words for TF data generation. It comprises a \textbf{taxonomy} of market factors that impact the fluctuation of treasury futures and a \textbf{structured language} that standardizes the prompt writing. As a taxonomy, FinMAP features a hierarchical architecture that captures market signals across two temporal resolutions: the Daily Snapshot and the Periodic Situation.

\subsubsection{Hierarchical Taxonomy}
To derive a daily snapshot of the market, we focus on seven types of factors:
\begin{itemize} 
\item \textbf{Liquidity}: Quantifies central bank interventions (OMO/MLF) and monitors interbank borrowing costs (FR007/DR001) to assess systemic funding stress. 
\item \textbf{Sentiment}: Evaluates the qualitative "trading tone," including risk-on/off shifts and safe-haven demand, often reflected in equity market correlations. 
\item \textbf{Rates Bonds}: Tracks the yield curve movements of sovereign benchmarks and the technical performance (volume, open interest) of lead futures contracts. 
\item \textbf{Credit Bonds}: Evaluates credit spread fluctuations and secondary market liquidity across various rating grades and bond types (e.g., Tier-2 Capital Bonds). 
\item \textbf{Derivatives}: Monitors Interest Rate Swaps (IRS) and volatility metrics (implied vs. realized) to capture long-term benchmark expectations. 
\item \textbf{External Linkages}: Records global spillovers from foreign exchange markets (USDCNH), U.S. Treasury yields, and commodity price actions. 
\item \textbf{Event-Driven Factors}: Documents discrete policy shifts, macroeconomic data releases, and the market's internal pricing of upcoming scheduled events. 
\end{itemize}

On the other hand, to review the market situation over extended horizons (e.g., a week, month, or quarter), FinMAP underscores eight macroscopic categories: 
\begin{itemize} 
\item \textbf{Economic Themes}: Identifies the dominant economic narrative (e.g., "Cooling Inflation") that dictates investor behavior throughout the period. 
\item \textbf{Economic Environment}: Assesses the structural economic cycle, monetary policy stances, and the international geopolitical environment. 
\item \textbf{Key Prices}: Defines the boundary constraints of the market, including contract amplitude and critical technical/psychological support and resistance levels. 
\item \textbf{Technical Trends}: Describes the geometric evolution of price action (e.g., W-bottom patterns) and monthly-scale momentum indicators like MACD. 
\item \textbf{Cyclical Factors}: Accounts for historical analogies and seasonal anomalies, such as quarter-end liquidity tightening or "Spring Fever" effects. 
\item \textbf{Events and Timeline}: Maps the chronological chain of catalysts, from the initial release of data to its eventual absorption into market prices. 
\item \textbf{Market Sentiment}: Tracks the emotional trajectory of the market (Initial, Mid-term, and Late-phase) as influenced by evolving supply-demand pressures. 
\item \textbf{Risk Analysis}: Profiles potential tail risks, identifying asymmetric upside/downside triggers and systemic shocks that could disrupt the prevailing trend. 
\end{itemize}

\subsubsection{Structured Language}
Based on the factors that trace the market dynamics, it is not hard to develop structured prompts for conditional generation. 
Example prompts are demonstrated in Table~\ref{tab:p1}, Table~\ref{tab:p2}, and Table~\ref{tab:p3}, with a horizon of one day, one week, and one month, respectively.
It is noteworthy that, when preparing the dataset, we utilize daily snapshots of market services as a foundation for summarizing periodical market dynamics. This process is detailed in Appendix~\ref{sec:a3}.



\subsection{Data Collection}
\label{sec:a3}
We aggregated treasury futures market data from official open-access platforms\footnote{\url{https://www.wind.com.cn/}} and synthesized conditioning inputs by extracting latent factors from daily market reviews sourced from the Wind Terminal and proprietary in-house databases\footnote{The daily market review is published the next day.}. 

\begin{figure}[!t]
\centering
\includegraphics[width=0.95\linewidth]{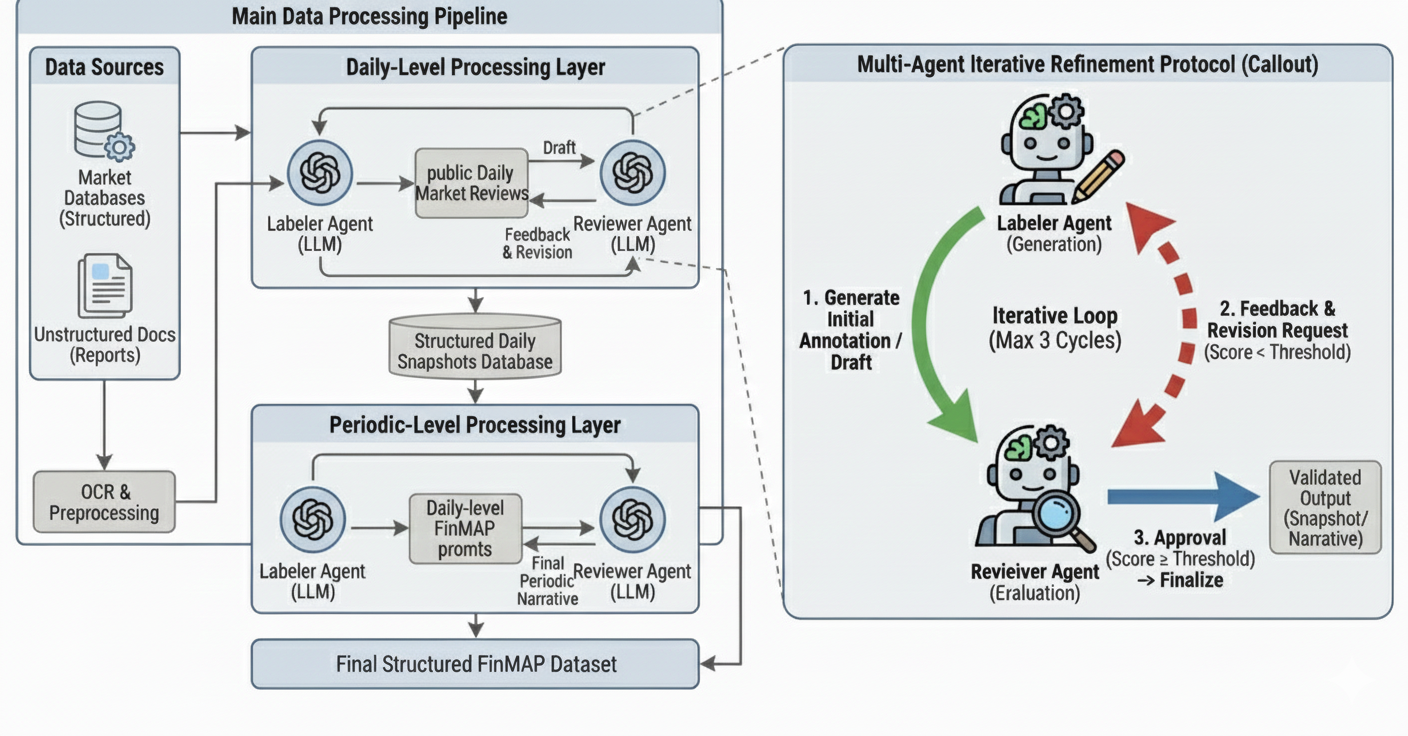}
\caption{A multi-agent framework for prompt labeling of treasury futures data.}
\label{fig:agent}
\end{figure}

To manage the high-throughput requirement of processing thousands of unstructured financial reports, we developed an automated labeling pipeline powered by a multi-agent collaborative framework. As illustrated in Figure~\ref{fig:agent}, the framework utilizes a Labeler-Reviewer architecture to ensure the fidelity of the extracted attributes. 
The process mirrors the hierarchical structure of the FinMAP protocol, executing in two distinct stages: 1) daily snapshot extraction, and 2) recursive periodic synthesis. At each stage, the agents engage in an iterative feedback loop, capped at three cycles, to achieve high-quality semantic grounding.

\subsubsection{Daily-Level Snapshot Extraction}
The daily labeling process follows a structured three-step protocol:
\begin{itemize}
\item \textbf{Pre-processing}: For each target date, raw market reviews are retrieved. Non-machine-readable formats (e.g., PDF reports) are pre-processed using Optical Character Recognition (OCR) to ensure text integrity before being passed to the Labeler agent.
\item \textbf{Structured Extraction}: The Labeler agent is tasked with parsing the review passages to identify and extract crucial economic factors defined by the FinMAP daily schema, outputting the result in a structured format.
\item \textbf{Iterative Quality Audit}: The Reviewer agent performs a quantitative assessment of the Labeler's output, assigning a score (1–5) based on its alignment with the source text. If the output receives a score below 4, the Reviewer generates specific revision comments. The Labeler then re-processes the prompt, incorporating this feedback to refine the extraction.
\end{itemize}
This cycle terminates when the Reviewer assigns a score of $\ge 4$ or the iteration limit of three loops is reached.

\subsubsection{Periodic Narrative Synthesis}
The period-level labeling leverages a hierarchical aggregation logic to capture long-term market regimes:
\begin{itemize}
\item \textbf{Recursive Data Retrieval}: Given a specified start date and temporal window, the pipeline retrieves previously generated snapshots. For weekly narratives, the system aggregates daily FinMAPs; for monthly narratives, it recursively aggregates weekly FinMAPs to maintain temporal coherence; for quarterly narratives, it recursively aggregates monthly FinMAPs, and so on.
\item \textbf{Macro-Factor Distillation}: The Labeler agent synthesizes the temporally ordered snapshots into a periodic narrative, focusing on macroscopic indicators such as economic themes and structural risk analysis.
\item \textbf{Consistency Review}: The Reviewer agent audits the periodic summary against the underlying snapshots to ensure no significant daily signals were lost in the synthesis. The scoring and feedback mechanism follows the same iterative logic as the daily-level extraction.
\end{itemize}

We implement the Labeler and Reviewer agents with Tongyi-DeepResearch~\footnote{\url{https://www.modelscope.cn/models/iic/Tongyi-DeepResearch-30B-A3B}} and Qwen3-235B~\footnote{\url{https://www.modelscope.cn/models/Qwen/Qwen3-235B-A22B-Thinking-2507}}, respectively. Section~\ref{sec:prompt} reports the prompts developed for these two agents.

\begin{table*}[!t]
\centering
\scriptsize
\caption{
Data distribution of our dataset.
\textit{Cont.}: contract categories.
}
\begin{tabular}{c|m{1cm}<{\centering}m{1cm}<{\centering}m{1cm}<{\centering}m{1cm}<{\centering}|p{1cm}<{\centering}p{1cm}<{\centering}p{1cm}<{\centering}p{1cm}<{\centering}}
\toprule
\multirow{2.5}{*}{\textbf{Cont.}} & \multicolumn{4}{c|}{\textbf{\# Training Samples}} &  \multicolumn{4}{c}{\textbf{\# Test Samples}} \\
\cmidrule{2-5}\cmidrule{6-9}
 & 8 & 32 & 64 & 128  & 8 & 32 & 64 & 128 \\
\midrule
TS & 1,582 & 1,558 & 1,526 & 1,462 & 193 & 169 & 137 & 73 \\
TF & 2,788 & 2,764 & 2,732 & 2,668 & 193 & 169 & 137 & 73 \\
T & 2,418 & 2,394 & 2,362 & 2,298 & 193 & 169 & 137 & 73 \\
TL & 448 & 424 & 392 & 328 & 193 & 169 & 137 & 73 \\
\midrule
Total & 7,236 & 7,140 & 7,012 & 6,756 & 772 & 676 & 548 & 292 \\
\bottomrule
\end{tabular}
\label{tab:data}
\end{table*}

\subsection{Datasets}
\label{sec:data}
\textbf{Processing}.
While treasury futures share similarities with equity markets (e.g., stock price), their price dynamics exhibit fundamental differences. Unlike the often non-stationary nature of stock prices, TF prices are bounded by the underlying bond's par value, typically oscillating within a stable range (e.g., 90 to 110) with significant mean-reverting characteristics. Consequently, traditional log-return normalization - while standard for equities - may fail to capture the structural stationarity of TF markets. 
We apply a stratified normalization strategy to three distinct feature groups. 
For price-based variables: opening price, highest price, lowest price, closing price, and settlement price, we normalize them based on the previous day's opening price ($Open_{t-1}$):
$$
X_t^{norm} = \frac{X_t - Open_{t-1}}{Open_{t-1}} \times 100
$$
This approach preserves the magnitude of relative fluctuations while centering the data to facilitate model convergence.
For volume and liquidity metrics, trading volume and trading value, we apply a log-transformation to compress the dynamic range and mitigate the impact of outliers:
$$
X^{norm} = \log_{10}(X + 1)
$$
For the open interest that captures the momentum of market participation and positioning shifts, we normalize it using the first-order growth rate:
$$
X_t^{norm} = \frac{X_t - X_{t-1}}{X_{t-1}}
$$

\textbf{Splitting}.
Our collected dataset comprises 2,625 prompt-time series pairs across four contracts, spanning from March 20, 2015, to December 31, 2025.
For each contract, we retain the data from the last 200 days for testing purposes, while the remaining data is utilized for model training. Table~\ref{tab:data} presents the data distributions based on the duration of synthesis tasks.

\subsection{Implementation Details}
\label{sec:a4}
All our experiments are conducted on 8 Nvidia H200 GPUs. Our codes are mainly implemented with Pytorch\footnote{\url{https://pytorch.org/}} and Huggingface Transformers\footnote{\url{https://huggingface.co}}. 
We train the U-VAE model using the AdamW optimizer~\cite{Adamw} with a learning rate of 0.001 and a cosine learning rate scheduler. Similarly, the TF-CoDiT model is trained with the same optimizer and scheduler, but with a learning rate of 0.0005 and a warming-up rate of 0.05.
The hyperparameters in terms of model architectures are reported in Table~\ref{tab:vae} and Table~\ref{tab:dit}. More details are as follows.

\begin{table}[!t]
\centering
\small
\caption{Hyperparameter specifications for U-VAE.}
\label{tab:vae}
\begin{tabular}{ccc}
\toprule
\textbf{Item} & \textbf{Encoder} & \textbf{Decoder} \\ 
\midrule
Hierarchical Layers ($L$) & 3 & 3 \\
Reduction Ratio ($r$) & 2 & 2 \\
Hidden Dimension ($d$) & 64 & 64 \\
Attention Heads ($h$) & $\{16, 8, 4\}$ & $\{4, 8, 16\}$ \\
\bottomrule
\end{tabular}
\end{table}

\begin{table}[!t]
\centering
\small
\caption{Architectural specifications of Gemma2.}
\label{tab:dit}
\begin{tabular}{cc}
\toprule
\textbf{Item} & \textbf{Value} \\
\midrule
Hidden Layers & 26 \\
Hidden Dimensions & 2,304 \\
Attention Heads & 8 \\
Projection Dimensions & 9,216 \\
Vocabulary Size & 256,000 \\
Context Size (Tokens) & 8,192 \\ 
Parameters (Billion) & $\sim 2.61$ \\
\bottomrule
\end{tabular}
\end{table}

\subsubsection{Model Training}
Because the U-VAE is a separate component of our TF-CoDiT, we trained U-VAE and TF-CoDiT (the LLM that performs conditional diffusion generation) using distinct data recipes and strategies.

\textbf{U-VAE}.
We train U-VAE to encode DWT coefficients of time series, especially their inter-channel correlation, into a latent variable. For this purpose, we curate the training set based on our in-hand TF data. We apply a sliding window approach with a size $S \in \{4, 8, 16, 32, 64, 96, 128\}$ and a stride of 1 on our four-contract TF time-series data, where each time series has a length of 2,625. This process yields 72,448 samples for the model training.

\textbf{TF-CoDiT}.
We train TF-CoDiT to generate the latent variable derived from the DWT coefficients that decomposes the time series. Specifically, TF-CoDiT starts from FuseDiT, which is based on a Gemma-2B backbone. We freeze this LLM while adding a novel predicting head and replacing the visual embedding layer with a novel one initialized randomly. 
Additionally, the 2D RoPE position embedding at each attention layer, which originally locates the pixels by latitude and longitude, is redirected to the channels by time and frequency.
We train TF-CoDiT using the training sets listed in Table~\ref{tab:data}, by mixing varying contracts and durations, with only the predicting head and visual embedding layer updated. 

\subsubsection{Comparison Models}
We compared TF-CoDiT in experiments with the following models: \\
\textbf{WGAN}: a generative model formulated for the optimal transport problem that uses the Wasserstein distance to provide a more stable training process and generate high-fidelity synthetic data. \\
\textbf{QuantGAN}: a GAN framework that utilizes temporal convolutional networks to capture long-range dependencies and volatility clusters in financial time series. \\
\textbf{TimeGAN}: a novel framework that preserves temporal dynamics by combining unsupervised adversarial training with a stepwise supervised loss and a learned embedding space. \\
\textbf{Time-LLM}: an autoregressive framework that adapts frozen large language models (LLMs) for time-series forecasting by reprogramming input data into text prototypes and using "Prompt-as-Prefix" to provide natural language context and task instructions. \\
\textbf{CALF}: a training framework for LLM, which aligns the distribution discrepancies between textual and temporal modalities through a multi-level cross-modal fine-tuning approach that includes input matching, feature regularization, and output consistency. \\
\textbf{FinDDPM}: utilizes Denoising Diffusion Probabilistic Models (DDPMs) to generate synthetic financial data by converting multiple time series into images via wavelet transformation and then reconstructing them through inverse transformation. \\
\textbf{GBMDiff}: a diffusion framework that explicitly incorporates Geometric Brownian Motion into the forward noising process of a score-based model to reflect the price-dependent heteroskedasticity inherent in financial markets. \\
\textbf{TIMEDIT}: a "proto-foundation" diffusion transformer that combines temporal dependency learning with a unified masking mechanism and a fine-tuning-free editing strategy to handle diverse tasks like forecasting, imputation, and generation. \\
\textbf{T2S}: a domain-agnostic diffusion framework that bridges natural language and time series by using a length-adaptive VAE and DiT-based Flow Matching to generate high-resolution sequences of arbitrary lengths.

In our experiments, we utilized the \textit{repro-gan} library to implement WGAN and QuantGAN, while TimeGAN was deployed using its official implementation. FinDDPM, GBMDiff, TIMEDIT, and T2S are reproduced based on the Huggingface Diffusers library. 
A significant architectural adjustment was made to FinDDPM: while the original framework is constrained to 3-channel inputs to align with standard RGB image formats, we extended the architecture to 4 channels to generate OHLC prices. 
For controllable models (all of them controlled using natural language), we truncate our FinMAP prompt to fit their context size. Conversely, for uncontrollable models, the generation process is initiated from a standard Gaussian noise distribution to synthesize either raw time series sequences or wavelet coefficient matrices.

\begin{table}[!t]
\centering
\scriptsize
\caption{
Comprehensive comparison of baselines.
\textit{GAN}: Generative Adversarial Network.
\textit{AR}: Autoregressive model.
\textit{DM}: Diffusion Model.
}
\begin{tabular}{c|c|c|c|c}
\toprule
\textbf{Model} & \textbf{Architecture} & \textbf{Controlled} & \textbf{Multimodal} & \textbf{Pretrained} \\
\midrule
WGAN & GAN & \ding{55} & \ding{55} & \ding{55} \\
QuantGAN & GAN & \ding{55} & \ding{55} & \ding{55} \\
TimeGAN & GAN & \ding{55} & \ding{55} & \ding{55} \\
Time-LLM & AR & \ding{51} & \ding{51} & \ding{51} \\
CALF & AR & \ding{51} & \ding{51} & \ding{51} \\
GBMDiff & DM & \ding{55} & \ding{55} & \ding{55} \\
FinDDPM & DM & \ding{55} & \ding{55} & \ding{55} \\
TimeDiT & LLM-DiT & \ding{51} & \ding{51} & \ding{51} \\
T2S & LLM-DiT & \ding{51} & \ding{51} & \ding{51} \\
TF-CoDiT & LLM-DiT & \ding{51} & \ding{51} & \ding{55} \\
\bottomrule
\end{tabular}
\label{tab:models}
\end{table}

\subsubsection{Evaluation}

In our experiments, we evaluate the models' performance in reconstructing the standard Open, High, Low, and Close (OHLC) price quartet for a given time span. 
We select these four variables as the primary targets because they constitute the fundamental price discovery signals utilized in technical analysis and risk management strategies. 
Accuracy in these benchmarks is critical for capturing intraday volatility and the structural boundaries of the market regime.
Quantitatively, we measure the divergence between the generated synthetic trajectories and the ground-truth observations using:
\begin{itemize}
\item \textbf{Mean Squared Error (MSE)}: $MSE$ penalizes significant outliers in price generation.
\item \textbf{Mean Absolute Error (MAE)}: $MAE$ provides a robust measure of the average linear error magnitude.
\end{itemize}

\newlist{tabitemize}{itemize}{1}
\setlist[tabitemize]{label=\tiny$\bullet$, leftmargin=*, nosep, before=\vspace{-0.6\baselineskip}, after=\vspace{-1\baselineskip}}

\begin{table*}[t]
\centering
\scriptsize
\renewcommand{\arraystretch}{1.5}
\begin{tabularx}{\textwidth}{@{} l l c p{4.5cm} X @{}}
\toprule
\textbf{Category} & \textbf{Items} & \textbf{Abbr.} & \textbf{Indicators} & \textbf{Details (Market Logic \& Function)} \\
\midrule

\multirow{3}{*}{\textbf{Liquidity}} & Central Bank Ops & CBO & Scale, interest rates, and net injection/drain of OMO and MLF. & Monitors price and volume signals of short-to-medium term liquidity tools; serves as the primary window for policy fine-tuning. \\
\cmidrule(lr){2-5}
 & Funding Rates & FR & Key rates including DR007, FR007, and R001. & Benchmarks for interbank borrowing costs. DR007 is the central policy target; its deviation reflects instantaneous liquidity stress. \\
\cmidrule(lr){2-5}
 & NCD Market & NCD & 1Y State-owned/Joint-stock NCD rates, issuance spreads, and net financing. & Core metric for commercial banks' medium-term liability costs; reflects bank-side liquidity demand and future interest rate expectations. \\
\midrule

\multirow{2}{*}{\textbf{Sentiment}} & Market Sentiment & MS & Bond leverage levels (repo volume), futures basis/calendar spreads, and primary auction results. & Quantitative synthesis of risk appetite. Leverage reflects trading heat, while auction results indicate the strength of real demand. \\
\cmidrule(lr){2-5}
 & Equity Correlation & EC & Correlation between CSI 300 and 10Y T-bond yields; turnover ratios; "Stock-Bond Seesaw" intensity. & Monitors cross-asset capital flows. Marginal shifts in relative value drive capital reallocation and impact bond demand. \\
\midrule

\multirow{3}{*}{\textbf{Rates Bonds}} & Cash Bond Trends & CBT & 10Y T-bond active yields, yield curve slope (10Y-1Y spread), and key level breakouts. & The pricing benchmark and trend core of the market. Curve morphology reflects economic outlooks and trend reversals. \\
\cmidrule(lr){2-5}
 & Futures Perf. & FP & Price and open interest (OI) of lead contracts, volume-to-OI ratio, and net basis (CTD). & Price discovery and sentiment amplifier. OI changes indicate capital direction, while net basis reflects arbitrage and delivery intent. \\
\cmidrule(lr){2-5}
 & Driving Factors & DF & Liquidity conditions, net bond supply, macro data, and policy expectations. & A systematic synthesis of multi-dimensional drivers used for attribution analysis and forward-looking judgment. \\
\midrule

\multirow{2}{*}{\textbf{Credit Bonds}} & Overall Perf. & OP & Credit spreads of AAA/AA+ notes; spread divergence between industrial and LGFV bonds. & Systematic risk pricing dashboard. Spread movements reflect credit environment tightening/easing; divergence shows risk appetite structure. \\
\cmidrule(lr){2-5}
 & Trading Char. & TC & High-yield trading ratio, valuation deviations, and primary buyer types (Wealth Mgmt, Funds). & Characterizes market microstructure. Valuation deviations reveal institutional positioning and the sustainability of trends. \\
\midrule

\multirow{2}{*}{\textbf{Derivatives}} & Int. Rate Swaps & IRS & 1Y/5Y Repo IRS rates and their spreads against Treasury yields. & Reflects expectations for future short-term funding rate averages; isolates pure interest rate expectations from liquidity premiums. \\
\cmidrule(lr){2-5}
 & Volatility & VOL & Implied volatility from bond options and historical yield volatility. & Quantifies market uncertainty. Implied volatility represents future pricing of risk, while historical volatility validates past realized fluctuations. \\
\midrule

\multirow{3}{*}{\textbf{External}} & FX Market & FX & USDCNH spot/midpoint deviation; 1Y NDF implied depreciation expectations. & Constraint indicator for cross-border capital flows and policy room. Exchange rate strength influences foreign demand for domestic bonds. \\
\cmidrule(lr){2-5}
 & Overseas Rates & OR & 10Y U.S. Treasury (UST) yields; 10Y China-U.S. yield spread. & The global risk-free rate anchor. UST yields exert external pressure on domestic rates via capital flows and exchange rate expectations. \\
\cmidrule(lr){2-5}
 & Precious Metals & PM & International gold prices (USD) and real interest rate (TIPS yield) fluctuations. & Mirror for global risk aversion and real rates. Significant gold price moves often corroborate global recession or haven expectations. \\
\midrule

\multirow{2}{*}{\textbf{Events}} & Major Events & ME & Domestic macro data releases, PBOC meetings, and global financial events. & A calendar of deterministic schedules and idiosyncratic risk sources likely to trigger market re-pricing or volatility. \\
\cmidrule(lr){2-5}
 & Expectation Mgmt & EM & Rate hike/cut probabilities implied by derivatives; market reaction before/after events. & Analyzes "priced-in" expectations versus actual outcomes to identify trading opportunities stemming from expectation gaps. \\
\bottomrule
\end{tabularx}
\caption{FinMAP Daily Market Attribute Taxonomy ($L_{day}$) with Key Indicators and Financial Logic.}
\label{tab:finmap_daily_full}
\end{table*}

\begin{table*}[t]
\centering
\scriptsize
\renewcommand{\arraystretch}{1.5}
\begin{tabularx}{\textwidth}{@{} l l c X @{}}
\toprule
\textbf{Categories} & \textbf{Items} & \textbf{Abbr.} & \textbf{Details (Description \& Indicators)} \\
\midrule

\textbf{Economic Theme} & Macro-Economic Theme & MET & A high-level synthesis and naming of the core macroeconomic drivers influencing the market during the observation phase. \\
\midrule

\multirow{4}{*}{\makecell[l]{\textbf{Economic}\\textbf{Environment}}} & Economic Period & EP & Identifies the current economic stage (e.g., Recession, Recovery) and its impact on asset allocation and capital flows. \\
\cmidrule(lr){2-4}
 & Monetary Policy & MP & Specific operations, adjustments, and policy rhetoric from the central bank regarding liquidity management, interest rates, and RRR. \\
\cmidrule(lr){2-4}
 & International Env. & IE & Market performance of major overseas economies (e.g., U.S. Treasury yields) and the transmission of global sentiment to the domestic market. \\
\cmidrule(lr){2-4}
 & External Policies & ETP & Monetary policy dynamics of major global central banks (e.g., Fed, ECB) and the resulting shifts in global liquidity. \\
\midrule

\multirow{3}{*}{\textbf{Key Prices}} & Open/Close Prices & OCP & Critical price points of the lead contract within the phase, used to characterize the price benchmark and overall amplitude. \\
\cmidrule(lr){2-4}
 & Support Price & SP & Significant price floors derived from technical indicators, historical lows, or psychological thresholds. \\
\cmidrule(lr){2-4}
 & Resistance Price & RP & Significant price ceilings derived from historical highs, upper technical boundaries, or psychological thresholds. \\
\midrule

\multirow{2}{*}{\makecell[l]{\textbf{Technical}\\textbf{Trends}}} & Moving Tendency & MT & Predictions of short-term price direction and targets based on chart patterns (e.g., W-bottoms, triangles). \\
\cmidrule(lr){2-4}
 & Periodic Lines & PL & Assessment of medium-to-long term trends based on weekly or monthly technical indicators (e.g., MACD, moving averages). \\
\midrule

\multirow{3}{*}{\makecell[l]{\textbf{Cyclical}\\textbf{Factors}}} & Hist. Same Period & HSP & Statistical average performance or rising probability for the same calendar period over previous years as a historical reference. \\
\cmidrule(lr){2-4}
 & Seasonal Effect & SE & Recurrent market fluctuation patterns caused by seasonal or periodic factors (e.g., quarter-ends, major holidays). \\
\cmidrule(lr){2-4}
 & Calendar Effect & CE & Periodic spikes in market volatility triggered by fixed schedules (e.g., data release dates, central bank meetings). \\
\midrule

\multirow{4}{*}{\makecell[l]{\textbf{Events \&}\\textbf{Timeline}}} & Crucial Events & EVT & Identification and scheduling of critical events occurring within the phase that are likely to impact the market. \\
\cmidrule(lr){2-4}
 & Events Development & ED & Actual progress, results, or specific data released regarding the identified crucial events. \\
\cmidrule(lr){2-4}
 & Expected Trend & ET & Prevailing market expectations or mainstream views regarding ongoing or upcoming key events. \\
\cmidrule(lr){2-4}
 & Event Impact & EI & Specific, quantifiable short-term effects of an event's actual outcome on market prices. \\
\midrule

\multirow{3}{*}{\makecell[l]{\textbf{Market}\\textbf{Sentiment}}} & Initial Sentiment & IS & The dominant market sentiment and its primary drivers at the start of the observation phase. \\
\cmidrule(lr){2-4}
 & Mid-term Sentiment & MS & Shifts in market sentiment and their corresponding triggers during the middle of the phase. \\
\cmidrule(lr){2-4}
 & End-term Sentiment & ES & The final state of market sentiment and its drivers at the conclusion of the phase. \\
\midrule

\multirow{3}{*}{\textbf{Risk Analysis}} & Upside Risks & UR & Potential bullish factors or scenarios that could trigger a structural upward breakout in market prices. \\
\cmidrule(lr){2-4}
 & Downside Risks & DR & Potential bearish factors or scenarios that could trigger a structural downward breakdown in market prices. \\
\cmidrule(lr){2-4}
 & Other Risks & OR & Potential risk events that may trigger severe market volatility beyond direct bullish or bearish directions. \\
\bottomrule
\end{tabularx}
\caption{The Proposed FinMAP Periodic Situation Taxonomy ($L_{phase}$) for Macro-Level Market Analysis.}
\label{tab:finmap_phase_schema}
\end{table*}

\subsection{Prompts}
\label{sec:prompt}

\begin{table*}
\begin{tcolorbox}[colback=gray!5,colframe=gray!50,title=Instructions for the Labeler to generate prompts that follow the daily-level FinMAP.]
\small
[Documents (public market daily review)] \\

You are a helpful financial assistant, excellent at information retrieval and summarization. There is a table that illustrates crucial factors that affect the movement of treasury futures: \\

[Daily-level Financial Market Attribute Protocol] \\

Please follow this table to identify crucial information from the given documents, and then write a novel market review like this example: \\

[An example of market review] \\

Start your writing: 
\end{tcolorbox}
\end{table*}

\begin{table*}
\begin{tcolorbox}[colback=gray!5,colframe=gray!50,title=Instructions for the Labeler to generate prompts that follow the period-level FinMAP.]
\small
[Documents (daily-level FinMAP prompts)] \\

You are a helpful financial assistant, excellent at information retrieval and summarization. There is a table that illustrates crucial factors that trace the movement of treasury futures during a period: \\

[Period-level Financial Market Attribute Protocol] \\

Please follow this table to collect the relevant information exhibited in the given documents, and then write a [weekly/monthly/quarterly] market review like this example: \\

[An example of market review] \\

Start your writing: 
\end{tcolorbox}
\end{table*}

\begin{table*}
\begin{tcolorbox}[colback=gray!5,colframe=gray!50,title=Instructions for the Reviewer to score the Labeler-generated prompts.]
\small







You are an expert financial analyst specializing in treasury futures. Please evaluate the quality of an AI-generated market review by comparing it against raw documents and the provided FinMAP Taxonomy. \\

\textbf{Raw Documents}: [Public market reviews] \\

\textbf{FinMAP Taxonomy}: [Daily/Period-level FinMAP, Table format] \\

\textbf{AI-Generated Market Review}: [Labeler-generated text] \\

Assess how effectively the generated prompt extracts and structures information according to the FinMAP schema. Provide a score from 1 to 5 for each of the following criteria: \\

\textbf{C1: Taxonomy Alignment}: Does the prompt accurately map information to the specific FinMAP categories (e.g., Liquidity, Technical Trends, Risk Analysis)? 
(\textit{5: Perfectly categorized; 1: Completely ignores the schema.}) \\

\textbf{C2: Information Density}: Does the prompt capture all key indicators mentioned in the review (e.g., specific OMO volumes, yield basis points, or specific event development)? 
(\textit{5: No critical data lost; 1: Highly superficial or missing key values.}) \\

\textbf{C3: Financial Logicality}: Does the prompt maintain the causal relationships described in the review (e.g., "Yields rose *because* of net liquidity drain")? 
(\textit{5: Causal logic preserved; 1: Causal links are broken or hallucinated.}) \\

Output your evaluations: \\

- Taxonomy Alignment Score: [1-5] \\
- Information Density Score: [1-5] \\
- Financial Logicality Score: [1-5] \\
- Revision Comments: [less than 5 sentences]
\end{tcolorbox}
\end{table*}

\begin{table*}[!t]
\small
\caption{
A sample of market review regularized by our daily-level FinMAP.
}
\begin{tabular}{|p{0.95\linewidth}|}
\toprule
\textbf{Market Review of Chinese Treasury Futures (On December 26, 2025)} \\
\hline
\textbf{Liquidity}
\begin{itemize}[leftmargin=*, nosep]
    \item \textbf{Central Bank Operations:} On December 25, the PBOC conducted a 177.1 billion CNY 7-day reverse repo (rate: 1.40\%) and a 400 billion CNY 1-year MLF operation (300 billion CNY MLF matured), achieving a net MLF injection of 100 billion CNY. This marked the 10th consecutive month of MLF net injections. The 2025 MLF net injection totaled 1161 billion CNY, compared to -1986 billion CNY in 2024.
    \item \textbf{Funding Rates:} Overnight rate fell 0.55bps to 1.2189\% (lowest since August 2025); 7-day rate rose 10.42bps to 1.4442\% (1-month high). DR007 weighted average remained below 1.26\%.
    \item \textbf{CD Market:} 1-month NCD rate increased 2.81bps to 1.85\% (1-month high), while 1-year NCD rates remained stable.
\end{itemize}

\textbf{Market Sentiment}
\begin{itemize}[leftmargin=*, nosep]
    \item \textbf{Risk Appetite:} Year-end trading activity remained muted. Risk sentiment was generally neutral, with a brief selloff in the afternoon due to market rumors ("small essay") followed by a rebound after clarification.
    \item \textbf{Equity Correlation:} Shanghai Composite fell 0.44\%, with over 4000 stocks declining across both exchanges. The Wind Real Estate Bond Index dropped 1.2\%.
\end{itemize}

\textbf{Interest Rate Bonds}
\begin{itemize}[leftmargin=*, nosep]
    \item \textbf{Bond Yields:} 10-year yield showed divergence, with long-term bonds (10Y+) weakening (10Y rose 0.5bp, 30Y rose 0-0.5bp).
    \item \textbf{Futures Performance:} Treasury bond futures (TL) closed down 0.09\%, with the 10-year main contract trading at 101.24\%.
    \item \textbf{Key Drivers:} Stable liquidity failed to drive yields lower, while year-end supply pressure weakened short-term bonds.
\end{itemize}

\textbf{Credit Bonds}
\begin{itemize}[leftmargin=*, nosep]
    \item \textbf{Overall Performance:} High-grade bonds remained stable, while lower-grade spreads widened. Vanke bonds underperformed ("22 Vanke 14", "22 Vanke 02", etc.). The Wind High-Yield Municipal Bond Index fell 0.01\%.
    \item \textbf{Trading Characteristics:} High-grade bonds traded at par, while lower-grade bonds traded 0.5-1bp below valuation.
\end{itemize}

\textbf{External Linkage}
\begin{itemize}[leftmargin=*, nosep]
    \item \textbf{FX Market:} Onshore CNY closed at 7.0050 (+5bps), while offshore CNY surged past 7.00 (15-month high). USD Index stabilized above 99.
    \item \textbf{Overseas Rates:} 2Y US Treasury rose 1.47bps, 10Y US Treasury fell 0.39bps.
    \item \textbf{Precious Metals:} Gold reversed gains to return to opening levels; silver traded sideways.
\end{itemize}

\textbf{Events}
\begin{itemize}[leftmargin=*, nosep]
    \item \textbf{Key Events:} 
    \begin{itemize}[leftmargin=*, nosep]
        \item MOF and 8 other ministries issued climate disclosure guidelines for enterprises (voluntary implementation).
        \item CSRC reported 2 trillion CNY in sci-tech innovation bonds issued since 2021.
    \end{itemize}
    \item \textbf{Expectation Management:} Market focused on PBOC liquidity operations and fiscal policy intensity. Expectations for improved liquidity post-holiday.
\end{itemize}

\textbf{Key Takeaways}
\begin{itemize}[leftmargin=*, nosep]
    \item Bond market remained range-bound, with stable liquidity but limited yield declines.
    \item Real estate bonds underperformed (index fell 1.2\%), while offshore CNY broke above 7.00.
    \item Attention shifted to PBOC's post-holiday liquidity operations and US CPI data.
\end{itemize}
\\
\bottomrule
\end{tabular}
\label{tab:p1}
\end{table*}

\begin{table*}[!t]
\small
\caption{
A sample of a weekly market review regularized by our period-level FinMAP.
}
\begin{tabular}{|p{0.95\linewidth}|}
\toprule
\textbf{Market Review of Chinese Treasury Futures (December 2–6, 2024)} \\
\hline
\textbf{Economic Theme} 
\begin{itemize}[leftmargin=*, nosep]
\item \textbf{Year-end liquidity gaming and policy expectation building.} Weak PMI below 50 reinforced weak recovery consensus. CPI stayed low, deflationary pressure persisted. Market focus shifted from policy magnitude to sustainability of cross-cycle monetary adjustment. RRR cut expectations diverged; MLF oversubscription widely expected.
\end{itemize}

\textbf{Economic Background}
\begin{itemize}[leftmargin=*, nosep]
    \item \textbf{Economic Cycle:} Policy-bottom to economic-bottom transition. Credit expansion below seasonal. Corporate medium-to-long term loan growth stable but not rebounding. Institutional allocation demand dominant.
    \item \textbf{Monetary Policy:} PBOC maintained minimal daily reverse repos (RMB 20--40bn), net weekly withdrawal of RMB 12bn. Q4 MPC meeting minutes (Dec 7) shifted wording from "strengthening counter-cyclical adjustment" to "precise and effective implementation," interpreted as lower RRR probability but stronger structural tools.
    \item \textbf{International Environment:} US 10Y yield ranged 4.2\%--4.3\%. China-US spread (10Y) deeply inverted at -130 to -140bp. Foreign outflow pressure marginally eased.
    \item \textbf{External Policy:} Eurozone inflation cooled, markets pricing ECB cut in Q1 2025. USD Index 103--104. USD/CNY near 7.25, neutral impact.
\end{itemize}

\textbf{Key Levels}
\begin{itemize}[leftmargin=*, nosep]
    \item \textbf{Start/End:} T2503 opened 102.85, closed 102.92 (+0.07, range 0.25\%). Active 10Y cash yield fell 1.5bp to 2.63\%.
    \item \textbf{Support:} Futures: 102.70 (20DMA), 102.55 (Nov consolidation low). Cash: 2.65\% psychological.
    \item \textbf{Resistance:} Futures: 103.00 handle, 103.15 (Nov high). Cash: 2.60\% strong resistance.
\end{itemize}

\textbf{Technical Trend}
\begin{itemize}[leftmargin=*, nosep]
    \item \textbf{Pattern:} Late-stage wedge consolidation on daily chart. Volume shrank from 38k to 21k lots. Open interest -3k to 185k lots, reflecting pre-policy window deleveraging.
    \item \textbf{Monthly:} Small bullish candle. MACD red bars narrowing, fast line nearing bearish crossover. Price remained above 10M and 20M MA, medium-term trend intact.
\end{itemize}

\textbf{Cyclical Factors}
\begin{itemize}[leftmargin=*, nosep]
    \item \textbf{Historical Seasonality:} Past 5 years (2019--2023) first week of December average +0.12\%, 60\% win rate, driven by year-end liquidity easing.
    \item \textbf{Seasonal Effects:} December faces triple funding pressure: bank year-end accounting, wealth management repatriation, government bond settlement. Money market rates typically rise 20--30bp, constraining long-end; short-end (1--3Y) outperforms on allocation demand.
    \item \textbf{Calendar Effects:} Early December pre-Central Economic Work Conference quiet period, highly sensitive to policy rumors on deficit ratio or special bonds.
\end{itemize}

\textbf{Timeline and Key Events}
\begin{itemize}[leftmargin=*, nosep]
    \item \textbf{Dec 2:} Caixin Services PMI 49.8 (prev 50.2), weaker than expected. Futures opened +0.05.
    \item \textbf{Dec 4:} Trade data: Nov exports -3.5\% YoY (USD). MLF: RMB 400bn matured, PBOC injected RMB 450bn (net +50bn) at 2.50\% unchanged. Market interpreted as "quantity-up, price-flat" neutral stance. Futures printed long upper shadow, long profit-taking evident.
    \item \textbf{Event Expectations:} Focus shifted to Dec 9 CPI/PPI release and mid-December Central Economic Work Conference. Consensus expects "stability with progress" tone, deficit ratio 3.2\%--3.5\%, below aggressive forecasts of 3.8\%.
    \item \textbf{Event Impact:} PBOC net withdrawal offset by MLF injection. DR007 rose from 1.75\% to 1.82\% but stayed near policy rate 1.80\%, limited bond impact.
\end{itemize}

\textbf{Market Sentiment}
\begin{itemize}[leftmargin=*, nosep]
    \item \textbf{Early (Dec 2--3):} Cautiously optimistic. Weak PMI reinforced "weak reality" narrative, but concerns over concentrated local government special bond issuance in December (~RMB 300bn quota) subdued position-taking.
    \item \textbf{Mid (Dec 4--5):} Wait-and-see. MLF outcome cooled RRR cut expectations, profit-taking emerged. Declining open interest reflected active long liquidation, but prop desks and rural commercial banks provided strong support above 2.65\% in cash bonds.
    \item \textbf{Late (Dec 6):} Marginally improved. Friday afternoon rumor of "PBOC symposium on RRR cuts over weekend" plus A-share decline drove safe-haven flows. T2503 closed near session high. Wind Bond Sentiment Index rebounded from 48 to 52.
\end{itemize}

\textbf{Risk Analysis}
\begin{itemize}[leftmargin=*, nosep]
    \item \textbf{Upside:} If Central Economic Work Conference sets 2025 deficit ratio above 3.5\% with ultra-long special bonds, plus 25bp RRR cut, 10Y yield breaks below 2.60\%, futures target 103.30 (cash 2.55\%).
    \item \textbf{Downside:} If CPI YoY >0.5\% or Fed dot plot shows only one 2025 cut, China-US spread widens beyond -150bp, foreign selling triggers 10Y yield rebound to 2.75\%, futures retreat to 102.40.
    \item \textbf{Other:} Year-end credit bond default risk (property sector) could spark liquidity crunch and indiscriminate rates selling. Surge in NCD issuance (1Y NCD rate 2.52\% to 2.58\%) may divert bond allocation funds. 
\end{itemize}\\
\bottomrule
\end{tabular}
\label{tab:p2}
\end{table*}

\begin{table*}[!t]
\scriptsize
\caption{
A sample of a monthly market review regularized by our period-level FinMAP.
}
\begin{tabular}{|p{0.95\linewidth}|}
\toprule
\textbf{Market Review of Chinese Treasury Futures (December 2–31, 2024)} \\
\hline


\textbf{Economic Themes}

\textbf{Escalating Supply Pressure}.  
The Ministry of Finance announced an intensified issuance schedule for ultra-long special treasury bonds in early December, with a focus on the 30-year tenor in Q4. Markets widely anticipated a significant increase in net supply. Primary market demand weakened, with bid-to-cover ratios falling to 2.1x, triggering a self-reinforcing feedback loop: “harder to sell $\rightarrow$ lower prices $\rightarrow$ greater investor reluctance.”

\textbf{Pronounced Policy Divergence}.  
On December 10, the U.S. Federal Reserve cut rates by 25 bps to 3.50\%–3.75\%, yet internal dissent was evident. Meanwhile, on December 15, the PBOC maintained the MLF rate unchanged while injecting only CNY 100 billion net liquidity—dashing market expectations for easing. This policy misalignment led to oscillating sentiment between caution and optimism.

\textbf{Credit Risk Flare-up}.  
On December 15, Vanke’s proposed extension for two medium-term notes totaling CNY 5.7 billion was rejected by bondholders. Although granted a 30-day grace period, the event intensified concerns over real estate sector solvency.

\textbf{Economic Background}

\textbf{Economic Cycle: Late Recession}  
The economy remains in the late stage of recession, with deteriorating corporate earnings. Some institutional investors began reducing long-duration holdings to manage NAV volatility, creating nonlinear selling pressure (“selling more as prices fall”).

\textbf{Monetary Policy: Moderately Accommodative but Constrained}  
In December, the PBOC conducted a net MLF injection of CNY 100 billion and purchased CNY 50 billion in government bonds—marking the third consecutive month of such operations. However, the unchanged MLF rate on December 15 fueled fears of an end to the “asset shortage” regime. On December 18, the PBOC Monetary Policy Committee signaled flexibility, stating it would “efficiently deploy a variety of tools, including RRR cuts and rate reductions”.

\textbf{International Environment: Weaker USD, Volatile UST Yields}  
Following the Fed’s rate cut, the U.S. dollar index weakened, accelerating foreign inflows into Chinese assets. However, the U.S. 10-year Treasury yield remained elevated at 4.127\%, sustaining a deep inversion in the China-U.S. yield spread of approximately –150 bps, limiting the appeal of RMB-denominated bonds.

\textbf{External Policy: Fed Dissent Amplifies Volatility}  
The Fed’s December meeting minutes revealed significant disagreement among officials regarding economic risks and the policy path forward. While the median FOMC projection for 2026 implied only one rate cut, market pricing anticipated ~50 bps of easing—widening the expectation gap and amplifying market swings.

\textbf{Key Price Levels}

\textbf{Opening/Closing Levels}
\begin{itemize}[leftmargin=*, nosep]
    \item Early December (Dec 1): T2603 opened at 108.33 and closed at 108.21; TL2603 opened near 113.89.
    \item Mid-December (Dec 12): T2603 closed at 107.985 (–0.21\% from Dec 1); TL2603 closed at 112.470 (–1.25\%).
\end{itemize}

\textbf{Support Levels}
\begin{itemize}[leftmargin=*, nosep]
    \item Technical: T2603 found short-term support near 107.60 (Dec 4 low); TL2603 broke below its key 112.80 support on Dec 4.
    \item Psychological: T2603 at 107.00; TL2603 at 112.00.
\end{itemize}

\textbf{Resistance Levels}
\begin{itemize}[leftmargin=*, nosep]
    \item Technical: T2603 faces resistance at 108.30 (prior high); TL2603 at 113.00.
    \item Psychological: T2603 at 108.50; TL2603 at 113.50.
\end{itemize}

\textbf{Technical Trends}

\textbf{Trend Evolution}
\begin{itemize}[leftmargin=*, nosep]
    \item T2603 formed a “V-shaped” rebound after the Dec 4 low of 107.61, recovering to 108.08 by Dec 18—but failed to breach the 108.30 resistance.
    \item TL2603 broke below 112.80 on Dec 4 and continued declining to 112.45, confirming bearish momentum on the long end.
\end{itemize}

\textbf{Monthly Chart Perspective}
\begin{itemize}[leftmargin=*, nosep]
    \item T2603 traded below the Bollinger Band midline in early December but moved back above it by Dec 12—yet remained capped by the upper band.
    \item TL2603 exhibited narrowing Bollinger Band width, reflecting sharply reduced risk appetite for ultra-long duration.
\end{itemize}

\textbf{Cyclical Factors and Historical Patterns}

\textbf{Historical Performance (First Half of December)}
\begin{itemize}[leftmargin=*, nosep]
    \item 2020: T2103 +0.7\%, 2021: T2203 +0.5\%, 2023: T2403 +0.2\%, 2024: T2412 +0.8\%
    \item \textbf{2025: T2603 –0.47\%}, significantly underperforming the historical average gain of 0.5\%–0.8\% and contrasting with the typical 70\%–80\% upside probability.
\end{itemize}

\textbf{Seasonal Effects}
\begin{itemize}[leftmargin=*, nosep]
    \item Historically, early December faces tax payments and year-end funding withdrawals, pushing DR007 higher.
    \item In 2025, DR007 stayed below 1.4\%, indicating markedly eased liquidity pressure due to sustained PBOC net injections.
\end{itemize}

\textbf{Calendar Effects}
\begin{itemize}[leftmargin=*, nosep]
    \item Key recurring dates—MLF operations around Dec 15 and PBOC committee meetings around Dec 18—aligned with 2025’s timeline.
    \item The broad selloff on Dec 4 resembled historical patterns during “data-dense windows” with insufficient policy communication.
\end{itemize}

\textbf{Timeline of Key Events}

\textbf{Key Events}
\begin{itemize}[leftmargin=*, nosep]
    \item Dec 4: TL2603 dropped to 113.670, a >1-year low, down >2.3\% from mid-November highs.
    \item Dec 10: Fed cut rates by 25 bps amid internal dissent.
    \item Dec 15: PBOC kept MLF rate unchanged (net CNY 100B injection); Vanke’s debt extension failed.
    \item Dec 18: PBOC signaled potential RRR/rate cuts.
    \item Dec 25: PBOC conducted an additional CNY 400B MLF operation (interpreted as year-end liquidity management).
\end{itemize}

\textbf{Event Developments \& Impact}
\begin{itemize}[leftmargin=*, nosep]
    \item Dec 4: TL2603’s plunge reflected concerns over long-term growth, fiscal sustainability, or rate trajectory.
    \item Dec 10: Fed cut weakened USD, boosting foreign inflow expectations—but domestic long-end yields remained pressured by supply.
    \item Dec 15: T2603 hit 107.61; TL2603 broke 112.80—driven by policy disappointment and credit shock.
    \item Dec 18: T2603 rebounded to 108.08 on PBOC’s dovish signal; TL2603 remained weak.
    \item Dec 25: T2603 showed limited recovery; TL2603 fell further to 112.51—highlighting the gap between policy expectations and reality.
\end{itemize}

\textbf{Market Expectations vs. Reality}
\begin{itemize}[leftmargin=*, nosep]
    \item Early Dec: Expected PBOC easing → bullish sentiment.
    \item Mid-Dec: Anticipated unchanged MLF rate + rising supply → bearish long end.
    \item Late Dec: Hoped for extra liquidity → range-bound outcome.
\end{itemize}

\textbf{Market Sentiment}

\begin{itemize}[leftmargin=*, nosep]
    \item \textbf{Early (Dec 1–3): Cautiously Optimistic} — Supported by PBOC liquidity injections and lack of negative data. Drivers: PBOC operations, data vacuum.
    \item \textbf{Mid (Dec 4–15): Turning Cautious} — Long-end weakness and “end of asset shortage” fears dominated. Credit concerns emerged but did not directly spill into rates. Drivers: supply, policy divergence, credit risk.
    \item \textbf{Late (Dec 18–25): Sentiment Recovery} — Short-end stabilized on PBOC signals; long-end remained depressed. Drivers: policy rhetoric, liquidity support.
\end{itemize}

\textbf{Risk Analysis}

\textbf{Upside Risks}
\begin{itemize}[leftmargin=*, nosep]
    \item Persistent economic underperformance could push T2603 above 108.50.
    \item Surprise PBOC rate/RRR cut could ignite bullish momentum.
\end{itemize}

\textbf{Downside Risks}
\begin{itemize}[leftmargin=*, nosep]
    \item Inflation rebound could drive T2603 below 107.50.
    \item Fed policy reversal strengthening USD could trigger foreign outflows.
\end{itemize}

\textbf{Other Risks}
\begin{itemize}[leftmargin=*, nosep]
    \item Credit contagion (e.g., Vanke default) causing liquidity shocks.
    \item Geopolitical escalation increasing rate volatility via safe-haven flows.
\end{itemize}
\\




\bottomrule
\end{tabular}
\label{tab:p3}
\end{table*}

\end{document}